\documentclass{article}

\PassOptionsToPackage{numbers, compress}{natbib}
\usepackage[final]{neurips_2024}

\usepackage[utf8]{inputenc} % allow utf-8 input
\usepackage[T1]{fontenc}    % use 8-bit T1 fonts
\usepackage{url}            % simple URL typesetting
\usepackage{booktabs}       % professional-quality tables
\usepackage{amsfonts}       % blackboard math symbols
\usepackage{nicefrac}       % compact symbols for 1/2, etc.
\usepackage{microtype}      % microtypography
\usepackage{xcolor}         % colors
\usepackage{graphicx}
\usepackage{wrapfig}
\usepackage{multirow}
\usepackage{algorithm2e}
\usepackage{xcolor,colortbl}
\definecolor{yellow}{rgb}{1, 1, 0.7}
\definecolor{orange}{rgb}{1, 0.85, 0.7}
\definecolor{red}{rgb}{1, 0.7, 0.7}

\title{Binocular-Guided 3D Gaussian Splatting with View Consistency for Sparse View Synthesis}
\author{%
  Liang Han$^1$, Junsheng Zhou$^1$, Yu-Shen Liu$^1$\thanks{The corresponding author is Yu-Shen Liu.} \ , Zhizhong Han$^2$ \vspace{0.15cm} \\
  School of Software, Tsinghua University, Beijing, China$^1$ \\
  Department of Computer Science, Wayne State University, Detroit, USA$^2$ \vspace{0.15cm} \\
  {\tt\small \{hanl23,zhou-js24\}@mails.tsinghua.edu.cn} \quad \\
  {\tt\small liuyushen@tsinghua.edu.cn \quad
  h312h@wayne.edu} \\
}

\begin{document}

\maketitle

\begin{abstract}
    Novel view synthesis from sparse inputs is a vital yet challenging task in 3D computer vision. Previous methods explore 3D Gaussian Splatting with neural priors (e.g. depth priors) as an additional supervision, demonstrating promising quality and efficiency compared to the NeRF based methods. However, the neural priors from 2D pretrained models are often noisy and blurry, which struggle to precisely guide the learning of radiance fields. In this paper, We propose a novel method for synthesizing novel views from sparse views with Gaussian Splatting that does not require external prior as supervision. Our key idea lies in exploring the self-supervisions inherent in the binocular stereo consistency between each pair of binocular images constructed with disparity-guided image warping. To this end, we additionally introduce a Gaussian opacity constraint which regularizes the Gaussian locations and avoids Gaussian redundancy for improving the robustness and efficiency of inferring 3D Gaussians from sparse views. Extensive experiments on the LLFF, DTU, and Blender datasets demonstrate that our method significantly outperforms the state-of-the-art methods. Project page is available at: \url{https://hanl2010.github.io/Binocular3DGS/}.
\end{abstract}

\section{Introduction}

3D reconstruction technologies \cite{NeRF, 3DGS} have demonstrated significant advances in synthesizing realistic novel views given a set of dense input views. To explore the challenging task in harsh real-world situations where only sparse inputs are available, some studies learn NeRF \cite{NeRF} with specially designed constraints \cite{dietnerf, sparsenerf, sparf, reconfusion} and regularizations \cite{regnerf, freenerf, diffusionerf} on the view scarcity. However, NeRF-based methods often suffer from slow training and inference speeds, leading to high computational costs that restrict their practical applications.

3D Gaussian Splatting (3DGS) \cite{3DGS} has achieved notable advantages in rendering quality and efficiency. However, 3DGS is still facing severe challenges with inputting sparse views, where the unstructured 3D Gaussians with limited constraints tend to overfit the given few views, resulting in geometric inaccuracies for scene learning. 
Some recent studies \cite{coherentgs, fsgs, dngaussian, sparsegs} on sparse view synthesis based on 3DGS employ the commonly-used depth priors from pre-trained models as additional constraints on the Gaussians geometries. However, the neural priors are often noisy and blurry, which struggle to precisely guide the learning of radiance fields. 

In this paper, we aim to design a method that does not require external prior as supervision, which directly explores the self-supervisions from the few input views for improving the quality and efficiency of sparse 3DGS. We justify that the key factors in achieving this goal include 1) learning more accurate scene geometry of Gaussians which leads to consistent views synthesis, and 2) avoiding redundant Gaussians near the surface for better efficiency and filtering noises.

For learning more accurate scene geometry of Gaussians, we explore the self-supervisions inherent in the binocular stereo consistency to constrain on the rendered depth of 3DGS solely utilized from given input views and synthesized novel views. Our key insight lies in the observation that binocular image pairs implicitly involve the property of view consistency, as demonstrated in the binocular stereo vision methods \cite{diggingbinocular, godard2017binocular, zhang2023lite-mono}. Specifically, we first translate the camera of one input view slightly to the left or right to obtain a translational view, from which we render the image and depth from 3DGS. The rendered image and the input one thus form a left-right view pair as in binocular stereo vision. We then leverage the rendered depth and the known camera intrinsic to compute the disparity of the view pair. We conduct the supervisions by warping the rendered image of the translational view to the viewpoint of the input image using the disparity, and constrains on the consistency between the warped and input images.

To further reduce redundant Gaussians near the scene surface and enhance the quality and efficiency of novel view synthesis, we propose a decay schema for the opacity of Gaussians. Specifically, we simply apply a decay coefficient to the opacity property of the Gaussians, penalizing the opacity during training.
To this end, Gaussians with lower opacity gradients (i.e., where the increase in opacity is smaller than the decay) are pruned, while Gaussians with higher opacity gradients (i.e., where the increase in opacity is greater than the decay) are retained. As the optimization process continues, redundant Gaussians are filtered out, and those newly generated (copied or split) Gaussians that are closer to the scene surface are retained, resulting in cleaner and more robust Gaussians.
This opacity decay strategy significantly reduces artifacts in novel views and decreases the number of Gaussians, improving both the rendering quality and optimization efficiency of 3DGS.

Additionally, to achieve better geometry initialization for improving 3DGS quality when conducting optimization on sparse views, we use pre-trained keypoints matching network \cite{pdcnet_plus} to generate a dense initialization point cloud. The dense point cloud describes the geometry of the scene more accurately, preventing Gaussians from appearing far from the scene surface, 
especially in low-texture areas where the distribution of Gaussians is subject to limited constraints.

In summary, our main contributions are as follows.
\begin{itemize}
    \item 
    We propose a novel method for synthesizing novel views from sparse views with Gaussian Splatting that does not require external prior as supervision.
    We explore the self-supervisions inherent in the binocular stereo consistency to constrain the rendered depth, solely obtained from existing input views and synthesized views. 
    \item We propose an opacity decay strategy which significantly regularizes the learning of Gaussians and reduces redundancy among Gaussians, leading to better rendering quality and optimization efficiency for novel view synthesis from sparse view with Gaussian Splatting.
    \item Extensive experiments on widely-used forward-facing and 360-degree scene datasets demonstrate that our method achieves state-of-the-art results compared to existing sparse novel view synthesis methods.
\end{itemize}

\section{Related Works}
\subsection{Neural Radiance Field}
Neural implicit functions have made great progress in surface reconstruction \cite{wang2021neus,Zhou2022CAP-UDF,huang2023neusurf,zhou2024fast,yariv2021volume,zhou2024cap,zhou2023levelset, zhang2024gspull}, 3D representation \cite{zhou2023uni3d,yan2023implicit,zhou20223d,li2023neaf,ma2023towards,Zhou2023VP2P,li2024LDI} and generation \cite{chou2023diffusion,udiff,zhou2024diffgs,wen20223d,zhou2024deep,ma2023geodream}. Detailed and realistic 3D scene representation has always been the research goal in the field of computer vision, and neural radiance fields (NeRFs) \cite{NeRF} has brought fundamental innovation to this domain. NeRF can reconstruct high-quality 3D scenes from sparse 2D images and generate realistic images from arbitrary viewpoints by representing scenes as continuous volume radiance functions. 

However, NeRF requires a large number of views as input during the training stage and exhibits limitations in both training and inference speed. Consequently, the following researches have primarily focused on addressing these bottlenecks by improving computational efficiency \cite{plenoxels, instant-ngp, dvgo, 3DGS, tri-mipnerf, tensorf, mvsnerf, k-planes} or reducing the number of input views \cite{regnerf, dietnerf, sparsenerf, sparf, geoaug, reconfusion, vipnerf, geconerf, depthnerf, darf, flipnerf, sun2023vgos, diffusionerf}, while continuously striving for enhanced rendering quality \cite{mipnerf, mipnerf360, zip-nerf}.

Notably, recent approach 3D Gaussian Splatting \cite{3DGS} has shown promising results in achieving real-time rendering capabilities without compromising rendering quality.

3D Gaussian Splatting \cite{3DGS} is an emerging method for novel view synthesis, which reconstructs high quality scenes rapidly by utilizing a set of 3D Gaussians to represent the radiance field in the scene. This method performs excellently when dealing with real-world scenes, particularly excelling in handling high-frequency details. Moreover, 3D Gaussian Splatting demonstrates significant advantages in inference speed and offers more intuitive editing and interpretability capabilities.

\subsection{Novel View Synthesis from Sparse Views}
Recent researches \cite{regnerf, dietnerf, sparsenerf, sparf, geoaug, reconfusion, vipnerf, geconerf, depthnerf, darf, flipnerf, sun2023vgos, diffusionerf} have explored various approaches to generate novel views from sparse input images, focusing on enhancing both rendering quality and efficiency. Methods such as NeRF\cite{NeRF} and 3DGS \cite{3DGS} have been refined through various techniques, resulting in continuous improvement in the quality of novel view synthesis. 

Using depth priors obtained from pre-trained networks \cite{depth_anything, transformer4depth, robust_depth} to supervise neural radiance fields is a widely used technique \cite{darf, sparsenerf}. Some methods \cite{regnerf, freenerf, diffusionerf, depthnerf} introduce regularization terms in NeRF to address the problem, such as frequency regularization \cite{freenerf} and ray entropy regularization \cite{infonerf}. Additionally, there are also some methods \cite{diffusionerf, reconfusion, dietnerf} that leverage pre-trained models such as Diffusion model to enhance the rendering quality of novel views. However, most methods incur high costs during training and inference. While some methods have improved inference efficiency through generalizable models \cite{mvsnerf, pixelnerf, xu2024murf} or by using voxel grids \cite{sun2023vgos}, they often sacrifice rendering quality.

Currently, with the advent of 3D Gaussian splatting, some methods use 3DGS for sparse view synthesis, such as FSGS \cite{fsgs}, SparseGS \cite{sparsegs}, DNGaussian \cite{dngaussian} and CoherentGS \cite{coherentgs}. These methods utilize depth priors obtained from pre-trained models \cite{transformer4depth} as supervision. Depth constraints enable the unstructured Gaussians to approximately distribute along the scene surface, thereby enhancing the quality of novel view images. However, the depth priors from pre-trained models often contain significant errors and cannot make the Gaussians distribute to optimal positions. 
In contrast, our method employs view consistency constraints based on binocular vision, resulting in more accurate depth information and thereby achieving a more optimal distribution of Gaussians.
Some concurrent studies \cite{wolf2024gs2mesh, safadoust2024self} also employed the concept of binocular stereo in 3DGS \cite{3DGS}, but they feed binocular images into a pre-trained network to obtain depth priors, instead of self-supervision.

\begin{figure}
  \centering
  \includegraphics[width=1.0\linewidth]{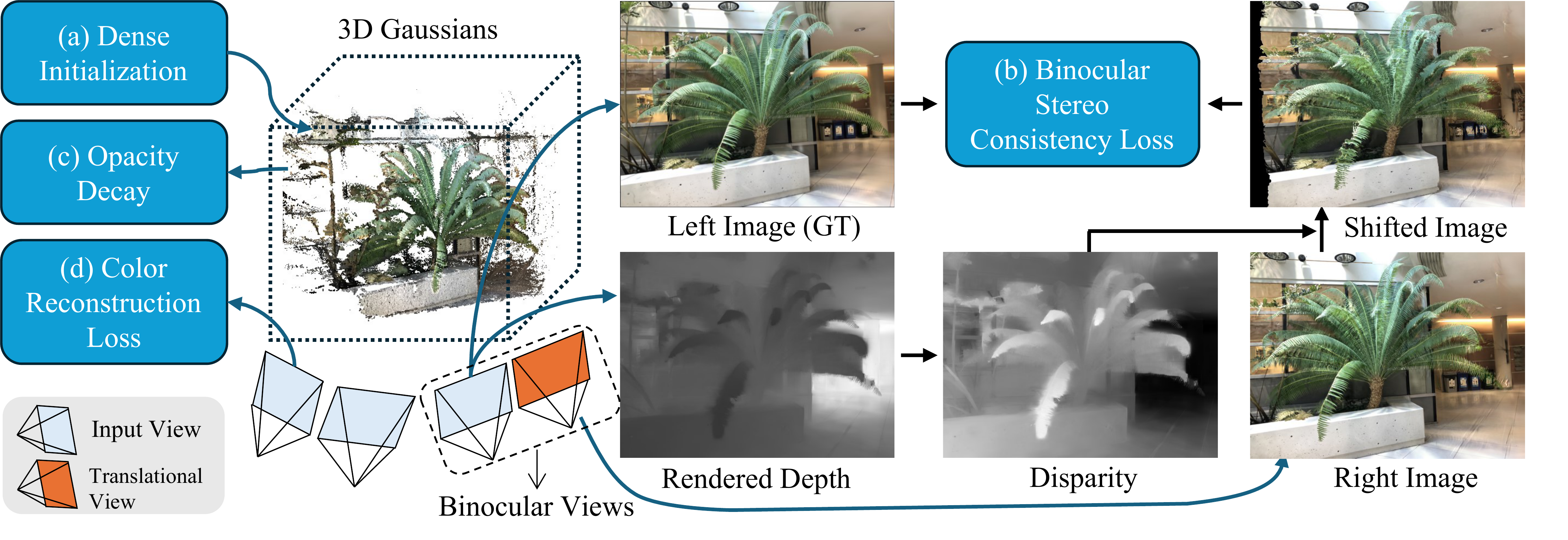}
  \caption{The overview of our method. (a) We leverage dense initialization for achieving Gaussian locations, and optimize the locations and Gaussian attributes with three constraints or strategies: (b) Binocular Stereo Consistency Loss. We construct a binocular view pair by translating an input view with camera positions, where we constrain on the view consistency of binocular view pairs in a self-supervised manner. (c) Opacity Decay Strategy is designed to decay the Gaussian opacity during training for regularizing them. (d) The Color Reconstruction Loss.}
  \label{pipeline}
\end{figure}

\section{Methods}
The pipeline of our method is depicted in Figure \ref{pipeline}. In this section, we first review the 3D representation method based on 3D Gaussians. Then, we explain how to construct stereo view pair and utilize it to enforce view consistency in a self-supervised manner. Next, we introduce the opacity decay strategy and the dense initialization method for 3D Gaussians. Finally, we present the overall loss function used for optimization.

\subsection{Review of 3D Gaussian Splatting}
Gaussian splatting \cite{3DGS} represents scene with a set of 3D Gaussians. Each 3D Gaussian is defined by a central location $\mu \in \mathbb{R}^3$, and a covariance matrix $\Sigma \in \mathbb{R}^{3\times3}$. Formally, it is defined as 
\begin{equation}
    G_i(x) = e^{-\frac{1}{2}(x-\mu_i)^T\Sigma^{-1}(x-\mu_i)},
\end{equation}
where the covariance matrix $\Sigma$ has physical meaning only when it is positive semi-definite. Therefore, it can be decomposed into $\Sigma=RSS^TR^T$, $S\in\mathbb{R}^3$ is a diagonal scaling matrix with 3 parameters, $R\in\mathbb{R}^4$ is a rotation matrix analytically expressed with quaternions. In addition, for rendering the image, each Gaussian also stores an opacity value $\alpha \in\mathbb{R}$ and a color feature $f\in\mathbb{R}^k$.

The 3D Gaussian is projected into the 2D image space when rendering an image, the projected 2D Gaussian is sorted by its depth value, and then alpha blending is used to calculate the color of each pixel,
\begin{equation}
    c = \sum_{i=1}^{n} c_i\alpha_i^{\prime} \prod_{j=1}^{i-1}(1-\alpha_j^{\prime}), 
\end{equation}
where $c_i$ is the color computed from feature $f$ and $\alpha^{\prime}$ is the opacity of the 2D Gaussian, which is obtained by multiplying the covariance $\Sigma^{\prime}$ of the 2D Gaussian by the opacity $\alpha$ of the corresponding 3D Gaussian. The 2D covariance matrix $\Sigma^{\prime}$ is calculated by $\Sigma^{\prime}=JW\Sigma W^TJ^T$,  where $J$ is the Jacobian of the affine approximation of the projective transformation. $W$ is the view transformation matrix.

The optimization process of Gaussian splatting is initialized with a set of 3D Gaussians from a sparse point cloud from SfM. Subsequently, the density of the Gaussian set is optimized and adaptively controlled. During optimization, a fast tile-based renderer is employed, allowing competitive training and inference times compared to the fast NeRF based methods.

\subsection{Binocular Stereo Consistency Constraint}

The key factor in improving the rendering quality of 3DGS is to guide the Gaussians to be distributed close to the exact scene surfaces. To this end, we aim to design a constraint on the rendered depth of 3DGS which directly represents the Gaussian geometry. Previous studies commonly adopt the depth priors from pretrained models to guide the depth of 3DGS. However, the depth priors are often noisy and blurry, especially for complex scenes, which fails in providing accurate guidance.

In this paper, we propose a novel prior-free method which leverages the binocular stereo consistency in 3D Gaussian splatting. We constrain on the rendered depth of 3DGS solely obtained by the existing input views and the novel views rendered by 3DGS. Specifically, we first render a novel view from a view point which is translated from one of the input views, leading to a pair of binocular vision images. We then shift the novel view to the perspective of the input image using disparity. Finally, we constrain the rendered image using the consistency between the input image and the warped image.

Specifically, given an image view {$I_l$} as input, we translate its corresponding camera position $C_l$ to the right by a distance {$d_{cam}$} and obtain the translational camera position $O_r$. We then obtain a novel rendered image {$I_r$} by rendering 3D Gaussians from $O_r$. The images {$I_l$} and {$I_r$} form a binocular stereo image pair, where {$I_l$} is the left image and {$I_r$} is the right one, respectively. Simultaneously, we can obtain the rendered depth {$D_l$} corresponding to the left image {$I_l$} by rendering 3DGS from $O_l$. According to the geometric theory of binocular stereo vision, the relationship between disparity $d$ and depth $D_l$ is given by $ d=f\cdot d_{cam}/D_l$, 
where {$d \in \mathbb{R}^{h\times w}$} and each value $d_{ij}$ in $d$ indicates the horizontal shift required to align a pixel in the right image with its counterpart in the left image. {$f$} is the focal length of the camera. Then, we can use the disparity {$d$} to move each pixel of the right image {$I_r$}, obtaining the reconstructed left image {$I_{shifted}$}, as denoted by 
\begin{equation}
    I_{shifted}[i, j] = I_r[i-d_i, j-d_j].
\end{equation}
In practice, to make the reconstructed left image {$I_{shifted}$} differentiable, we use a bilinear sampler to interpolate on the right image {$I_r$} to obtain each pixel of {$I_{shifted}$}. The sampling coordinates for each pixel are directly obtained from the disparity.
Formally, we define the loss function by
\begin{equation}
\label{equ:L_consis}
    L_{consis}=\frac{1}{N}\sum_{i,j}\left | I_{ij}^{l}-I_{ij}^{shifted} \right |. 
\end{equation}

To achieve better optimization with the loss function, we do not use a fixed camera position shift in practice. Instead, we randomly translate the input camera to the left or right by a distance within a range [$-d_{max},d_{max}$] in each iteration.

Although some methods \cite{darf, geoaug, geconerf} use unseen views or neighboring training views as source views and warp them to the reference view for consistency constraints, they do not account for the impact of the distance and angle between the source and reference views on the effectiveness of the supervision. When the source view is rotated or moved away from the reference view, the warped image suffers severe distortion due to depth errors and occlusions. Compared to the ground truth (GT) image, this leads to larger errors, hindering the convergence of the image-warping loss. In contrast, slight camera translations that cause minor changes in view without rotation and negligible impact from occlusions, allow the errors between the warped image and the GT image to primarily stem from depth errors, facilitating better optimization of depth. For the results using different source views, please refer to the Appendix.

\subsection{Opacity Decay Strategy}
We further justify that relying solely on the depth constraints does not always lead to correct Gaussian geometries that are closely aligned with the exact scene surfaces. The reason is that the rendered depth varies with changes in the scale and opacity of the Gaussians, rather than being solely determined by their positions. While 3DGS flexibly optimizes the scale and opacity during training, which leads to deviations in the novel views.
To address this issue, we design a simple strategy by applying a decay coefficient {$\lambda $} to the opacity $\alpha$ of the Gaussians, penalizing the opacity during training.
\begin{equation}
    \hat{\alpha} = \lambda \alpha, 0<\lambda <1.
\end{equation}

\begin{wrapfigure}{r}{0.4\linewidth}
    \centering
    \includegraphics[width=1.0\linewidth]{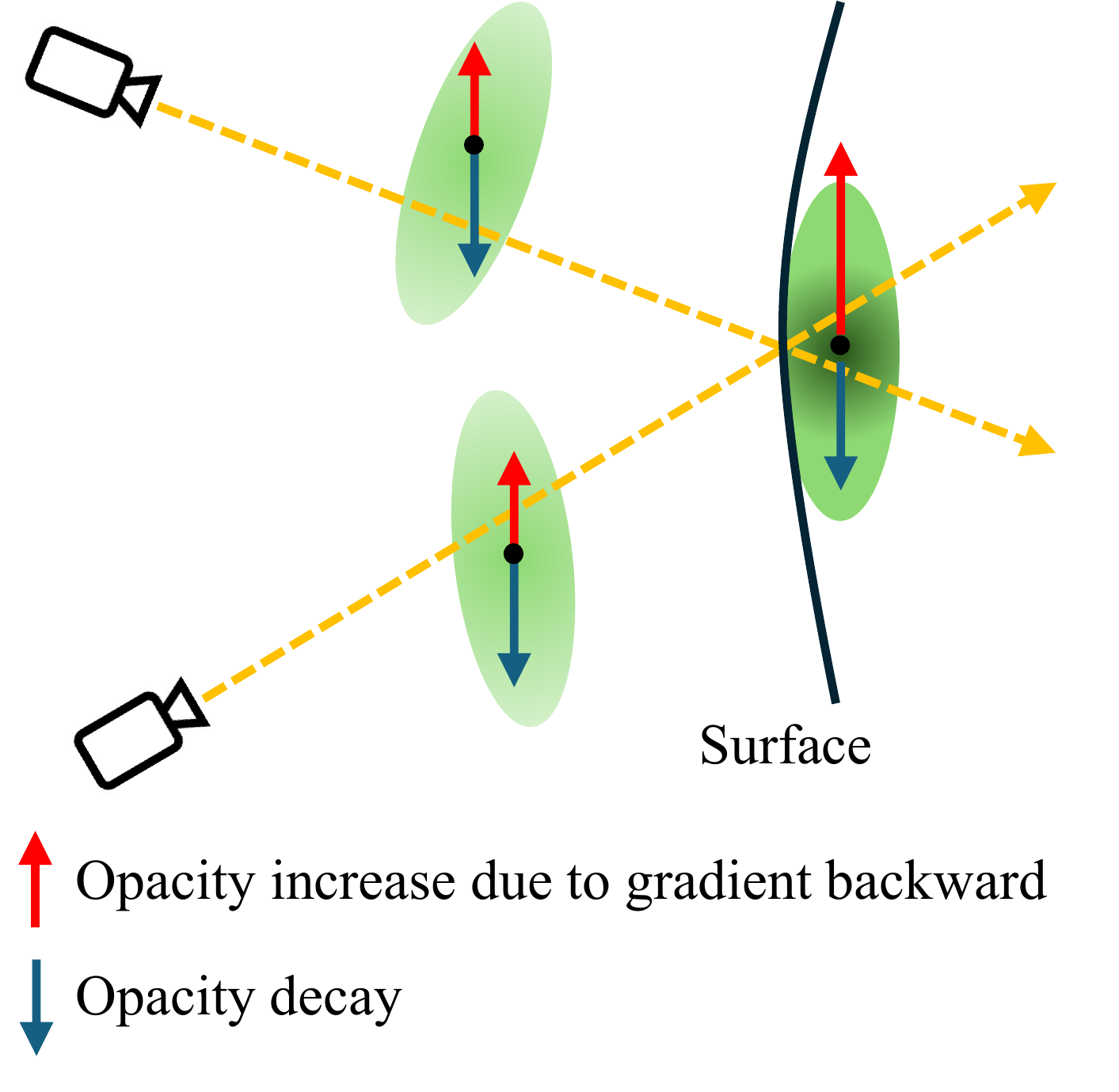}
    \vspace{-0.5cm}
    \caption{Illustration of the Gaussian opacity decay strategy.}
    \vspace{-0.5cm}
    \label{fig:opacity_decay}
\end{wrapfigure}

We illustrate the opacity decay strategy in Figure \ref{fig:opacity_decay}. 
Assuming all Gaussians are in the initialized state, due to the accumulation of constraints from multiple views, Gaussians near the scene surface typically have larger opacity gradients, allowing opacity to rise rapidly and construct the scene surface. However, some Gaussians far from the surface fail to get their opacity decreased and be pruned due to insufficient multiview consistent constraints, eventually affecting the rendering quality of novel views.

We aim to improve and stabilize the optimization of 3DGS by filtering out the far away Gaussians which indicates incorrect geometry while remain the ones close to the exact scene surfaces. This is achieved by applying the opacity decay strategy. We justify that the strategy does not lead all Gaussians' opacity going down, such as the ones on the surface, since these Gaussians' opacity progressively increase. As illustrated in Figure \ref{fig:opacity_decay}, the Gaussians far from the scene surface have lower opacity gradients due to fewer constraints, meaning their opacity increases less than those of the Gaussians on the scene surface. Under the opacity decay strategy, the opacity of Gaussians with lower opacity gradients gradually decreases until they are pruned. Conversely, the increase in opacity for Gaussians near the scene surface exceeds the decay magnitude, ultimately achieving a balance between the opacity increase and the decay, thereby preserving Gaussians close to the surface.

\begin{table}[]
\caption{Quantitative comparison on LLFF. We evaluate the NeRF-based and the 3DGS-based methods, our method achieves the best results in all metrics under different input-view settings.}
\label{tab:llff}
\resizebox{\linewidth}{!}{
\begin{tabular}{lccccccccc}
\toprule
         & \multicolumn{3}{c}{PSNR$\uparrow$} & \multicolumn{3}{c}{SSIM$\uparrow$} & \multicolumn{3}{c}{LPIPS$\downarrow$} \\
        Methods     & 3-view & 6-view & 9-view & 3-view & 6-view & 9-view & 3-view  & 6-view & 9-view \\
\midrule
        DietNeRF \cite{dietnerf}    & 14.94  & 21.75  & 24.28  & 0.370  & 0.717  & 0.801  & 0.496   & 0.248  & 0.183  \\
        RegNeRF \cite{regnerf}     & 19.08  & 23.10  & 24.86  & 0.587  & 0.760  & 0.820  & 0.336   & 0.206  & 0.161  \\
        FreeNeRF \cite{freenerf}   & 19.63  & 23.73  & 25.13  & 0.612  & 0.779  & 0.827  & 0.308   & 0.195  & 0.160  \\
        SparseNeRF \cite{sparsenerf}  & 19.86  & 23.26 & 24.27 & 0.714  & 0.741       & 0.781       & \cellcolor{yellow}0.243   & 0.235       & 0.228       \\
        ReconFusion \cite{reconfusion} & \cellcolor{orange}21.34  & \cellcolor{orange}24.25  & \cellcolor{yellow}25.21  & \cellcolor{orange}0.724  & \cellcolor{orange}0.815  & \cellcolor{yellow}0.848  & \cellcolor{orange}0.203   & \cellcolor{orange}0.152  & \cellcolor{orange}0.134  \\
        MuRF \cite{xu2024murf} & \cellcolor{yellow}21.26  & 23.54  & 24.66  & \cellcolor{yellow}0.722  & 0.796  & 0.836  & 0.245   & 0.199  & 0.164 \\
\midrule
        3DGS \cite{3DGS} & 15.52  &  19.45 & 21.13  & 0.405  & 0.627  & 0.715  & 0.408  & 0.268  & 0.214 \\
        FSGS \cite{fsgs} & 20.31  & \cellcolor{yellow}24.20  & \cellcolor{orange}25.32  & 0.652  & \cellcolor{yellow}0.811  & \cellcolor{orange}0.856  & 0.288   & \cellcolor{yellow}0.173  & \cellcolor{yellow}0.136  \\ 
        DNGaussian \cite{dngaussian}  & 19.12  & 22.18  & 23.17  & 0.591  & 0.755  & 0.788  & 0.294   & 0.198  & 0.180  \\
        Ours        & \cellcolor{red}21.44  & \cellcolor{red}24.87  & \cellcolor{red}26.17  & \cellcolor{red}0.751  & \cellcolor{red}0.845  & \cellcolor{red}0.877  & \cellcolor{red}0.168   & \cellcolor{red}0.106  & \cellcolor{red}0.090  \\
\bottomrule
\end{tabular}}
\end{table}

\begin{figure}[]
    \centering
    \includegraphics[width=1.0\linewidth]{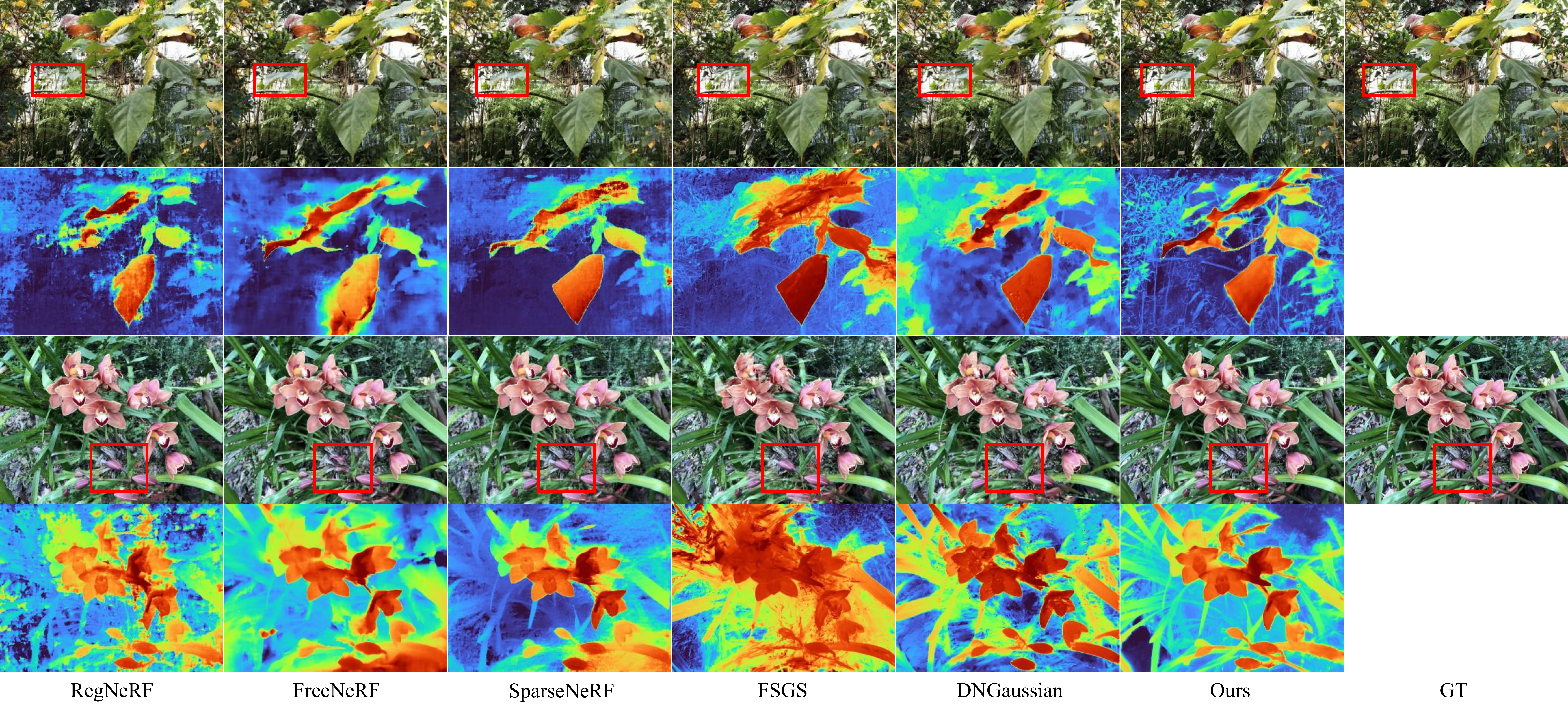}
    \caption{Visual comparison on LLFF dataset.}
    \label{fig:llff_visual}
\end{figure}

\subsection{Initialization from dense point clouds}
Previous 3DGS methods \cite{3DGS, fsgs, sparsegs} usually utilize a sparse point cloud generated by Structure from Motion (SfM) \cite{sfm} to initialize 3D Gaussians. However, the point cloud produced by sparse views is too sparse to adequately describe the scene to be reconstructed.
Although the splitting strategy in 3DGS can replicate new Gaussians to cover the under-reconstructed area, they are subject to limited geometric constraints and cannot adhere well to the scene surfaces, especially for low-texture areas where the distribution of Gaussians may be arbitrary. We therefore seek a robust approach to achieve better geometry initialization for improving 3DGS quality when optimizing from sparse views. Note that this prior is just used for initialization, and we do not use any this kind of priors during learning 3D Gaussians.

To achieve this, we use a pre-trained keypoints matching network \cite{pdcnet_plus} to generate a dense initialization point cloud. Specifically, we arbitrarily select two images from the input images, input them into the matching network, and obtain matching points. We then leverage the triangulation method \cite{triangulation}, along with the camera parameters corresponding to these images, to project the matching points into 3D space. This forms a dense point cloud, providing a more robust initialization for the Gaussians.

Compared with the sparse point cloud, the dense point cloud describes the geometry of the scene more accurately, preventing Gaussians from appearing far from the scene surface and ultimately leading to improved quality in novel view synthesis.

\subsection{Training Loss}
The final loss function consists of two parts: the proposed binocular stereo consistency loss {$L_{consis}$} as introduced in Eq.~(\ref{equ:L_consis}) and the commonly-used color reconstruction loss {$L_{color}$} of 3DGS \cite{3DGS}. We define the overall loss function by
\begin{equation}
    L = L_{consis} + L_{color},
\end{equation}
where {$L_{color}$} is composed of an {$L_1$} loss and a structural similarity loss {$L_{D-SSIM}$}, as denoted by
\begin{equation}
\label{equ:L_color}
    L_{color} = (1-\beta)L_1 + \beta L_{D-SSIM}.
\end{equation}

\begin{table}[]
\caption{Quantitative comparison on DTU. We evaluate the NeRF-based and the 3DGS-based methods, our method achieves the best results in most metrics under different input-view settings.}
\label{tab:dtu}
\resizebox{\linewidth}{!}{
\begin{tabular}{lccccccccc}
\toprule
         & \multicolumn{3}{c}{PSNR$\uparrow$} & \multicolumn{3}{c}{SSIM$\uparrow$} & \multicolumn{3}{c}{LPIPS$\downarrow$} \\
        Methods     & 3-view & 6-view & 9-view & 3-view & 6-view & 9-view & 3-view  & 6-view & 9-view \\
\midrule

        DietNeRF \cite{dietnerf}   & 11.85  & 20.63  & 23.83  & 0.633  & 0.778  & 0.823  & 0.214   & 0.201  & 0.173  \\
        RegNeRF \cite{regnerf}    & 18.89  & 22.20  & 24.93  & 0.745  & 0.841  & 0.884  & 0.190   & 0.117  & 0.089  \\
        FreeNeRF \cite{freenerf}   & 19.52  & 23.25  & \cellcolor{orange}25.38  & 0.787  & 0.844  & 0.888  & 0.173   & 0.131  & 0.102  \\
        SparseNeRF \cite{sparsenerf} & 19.47  &   -     &    -    & 0.829  &    -    &    -    & 0.183   &   -     &  -      \\
        ReconFusion \cite{reconfusion} & \cellcolor{orange}20.74  & \cellcolor{yellow}23.62  & 24.62  & \cellcolor{orange}0.875  & \cellcolor{yellow}0.904  & \cellcolor{yellow}0.921  & \cellcolor{orange}0.124   & \cellcolor{yellow}0.105  & \cellcolor{yellow}0.094  \\
        MuRF \cite{xu2024murf} & \cellcolor{red}21.31  & \cellcolor{orange}23.74  & \cellcolor{yellow}25.28  & \cellcolor{red}0.885  & \cellcolor{red}0.921  & \cellcolor{orange}0.936  & \cellcolor{yellow}0.127   & \cellcolor{orange}0.095  & \cellcolor{orange}0.084  \\
\midrule
        3DGS \cite{3DGS}       &  10.99   &  20.33   &   22.90   &  0.585    & 0.776         & 0.816    & 0.313      & 0.223       &  0.173      \\
        FSGS \cite{fsgs}       & 17.34  & 21.55  & 24.33  & 0.818  & 0.880  & 0.911  & 0.169   & 0.127  & 0.106  \\
        DNGaussian \cite{dngaussian} & 18.91  & 22.10  & 23.94  & 0.790  & 0.851  & 0.887  & 0.176   & 0.148  & 0.131  \\
        Ours        & \cellcolor{yellow}20.71  & \cellcolor{red}24.31  & \cellcolor{red}26.70  & \cellcolor{yellow}0.862  & \cellcolor{orange}0.917  & \cellcolor{red}0.947  & \cellcolor{red}0.111   & \cellcolor{red}0.073  & \cellcolor{red}0.052 \\
\bottomrule
\end{tabular}}
\end{table}

\begin{figure}[]
    \centering
    \includegraphics[width=1.0\linewidth]{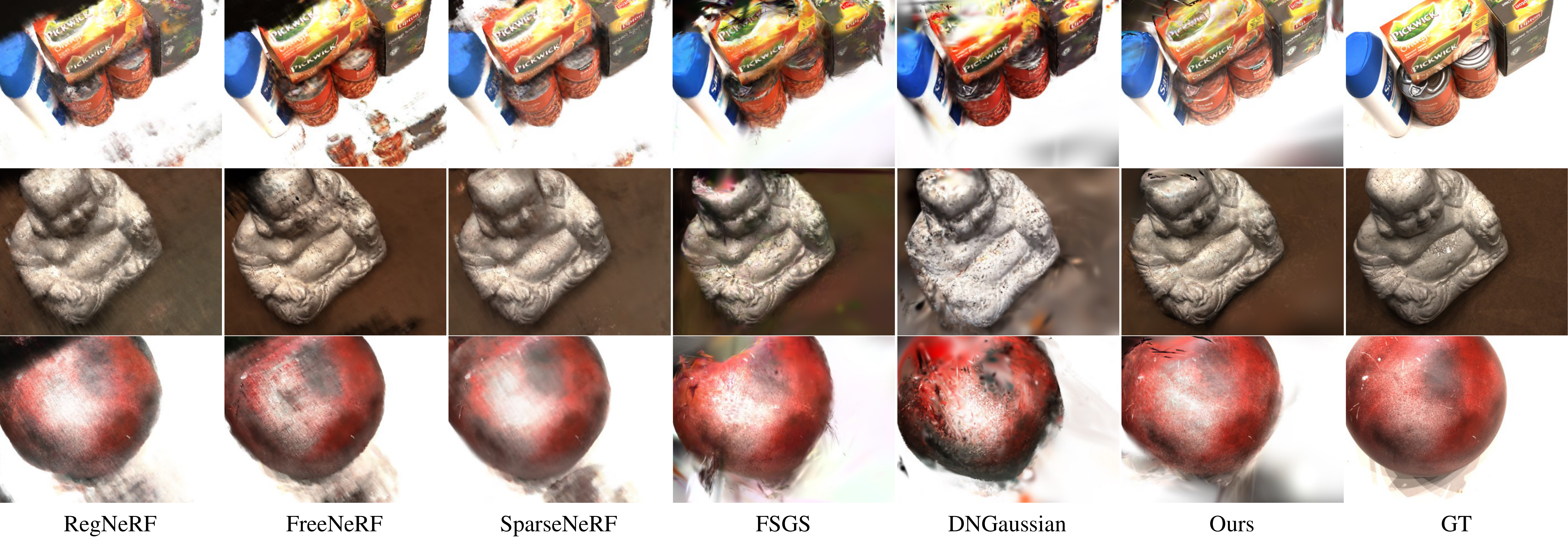}
    \caption{Visual comparison on DTU dataset.}
    \label{fig:dtu_visual}
\end{figure}

\section{Experiments}

\subsection{Datasets}
We conduct experiments on three public datasets, including the LLFF dataset \cite{llff}, the DTU dataset \cite{dtu} and the NeRF Blender Synthetic dataset (Blender) \cite{NeRF}. Following prior works \cite{regnerf, freenerf, dietnerf}, we used 3, 6, and 9 views as training sets for the LLFF and DTU datasets, and 8 images for training on the Blender dataset. The selection of test images remained consistent with previous works \cite{dietnerf, regnerf, freenerf}. The downsampling rates for the LLFF, DTU, and Blender datasets are 8, 4, and 2, respectively.

\subsection{Implementation details}
Since the LLFF and DTU are datasets of forward-facing scenes, they cannot be constrained by views from other directions during the optimization process. On the other hand, Blender is a dataset of 360-degree scenes, 8 images from different viewpoints are used as input during training, providing stronger constraints. Therefore, for the LLFF and DTU datasets, we utilize a pre-trained matching network PDC-Net+ \cite{pdcnet_plus} to obtain keypoints from input images, which are then used as the dense initialization point clouds. For the Blender dataset, we adopt random initialization as in the original 3DGS \cite{3DGS}.

\begin{wraptable}{r}{0.5\linewidth}
    \vspace{-0.5cm}
    \caption{Quantitative comparison on Blender for 8 input views. We evaluate the NeRF-based and the 3DGS-based methods, our method achieves comparable results to the state of the art methods.}
    \label{tab:blender}
    \centering
    \begin{tabular}{lccc}
    \toprule
         Methods  &  PSNR$\uparrow$ & SSIM$\uparrow$ & LPIPS$\downarrow$ \\
         DietNeRF \cite{dietnerf} & 22.50 & 0.823 & 0.124 \\
         RegNeRF \cite{regnerf} & 23.86 & 0.852 & 0.105 \\
         FreeNeRF \cite{freenerf} & \cellcolor{yellow}24.26 & \cellcolor{orange}0.883 & \cellcolor{orange}0.098 \\
         SparseNeRF \cite{sparsenerf} & 22.41 & 0.861 & 0.199 \\
    \midrule
         3DGS \cite{3DGS}    & 22.23 & 0.858 & 0.114 \\
         FSGS \cite{fsgs}    & 22.76 & 0.829 & 0.157 \\ 
         DNGaussian \cite{dngaussian} & \cellcolor{orange}24.31 & \cellcolor{red}0.886 & \cellcolor{red}0.088 \\
         Ours & \cellcolor{red}24.71 & \cellcolor{yellow}0.872 & \cellcolor{yellow}0.101 \\
    \bottomrule
    \end{tabular}
    \vspace{-0.5cm}
\end{wraptable}

Moreover, based on the analysis above, we train the LLFF and DTU datasets for 30,000 iterations, while the Blender dataset is trained for 7,000 iterations. Since the view consistency constraint based on binocular stereo vision needs to be performed on the basis of the training views that can already be rendered with high quality, we add the view consistency loss at 20,000 iterations for the LLFF and DTU datasets, and at 4,000 iterations for the Blender dataset. The maximum distance $d_{max}$ for camera shift is set to 0.4, the opacity decay coefficient $\lambda$ is set to 0.995, and the $\beta$ in the loss function \ref{equ:L_color} is set to 0.2 as in the original 3DGS \cite{3DGS}.

\subsection{Baseline}

We choose some state-of-the-art NeRF-based and 3DGS-based sparse view synthesis methods for comparison. NeRF-based methods include DietNeRF \cite{dietnerf}, RegNeRF \cite{regnerf}, FreeNeRF \cite{freenerf}, SparseNeRF \cite{sparsenerf}, ReconFusion \cite{reconfusion} and MuRF \cite{xu2024murf}. 3DGS-based methods include DNGaussian \cite{dngaussian} and FSGS \cite{fsgs}. Additionally, we compare against the original 3DGS \cite{3DGS}.

\subsection{Comparisons}

\paragraph{LLFF. } Table \ref{tab:llff} shows the quantitative results of the LLFF dataset with 3, 6, and 9 input views, respectively. Our method surpasses all baseline methods in terms of PSNR, SSIM \cite{ssim}, and LPIPS \cite{lpips} scores under different number of input views. When the number of input views increases to 9, it is almost adequate to provide sufficient color constraints. However, the evaluation scores of DNGaussian \cite{dngaussian} do not show a significant improvement, indicating that it does not perform well with an increased number of input views. This is because errors in the depth prior negatively affect the optimization process.

\begin{wrapfigure}{r}{0.5\linewidth}
    \centering
    \vspace{-0.5cm}
    \includegraphics[width=1.0\linewidth]{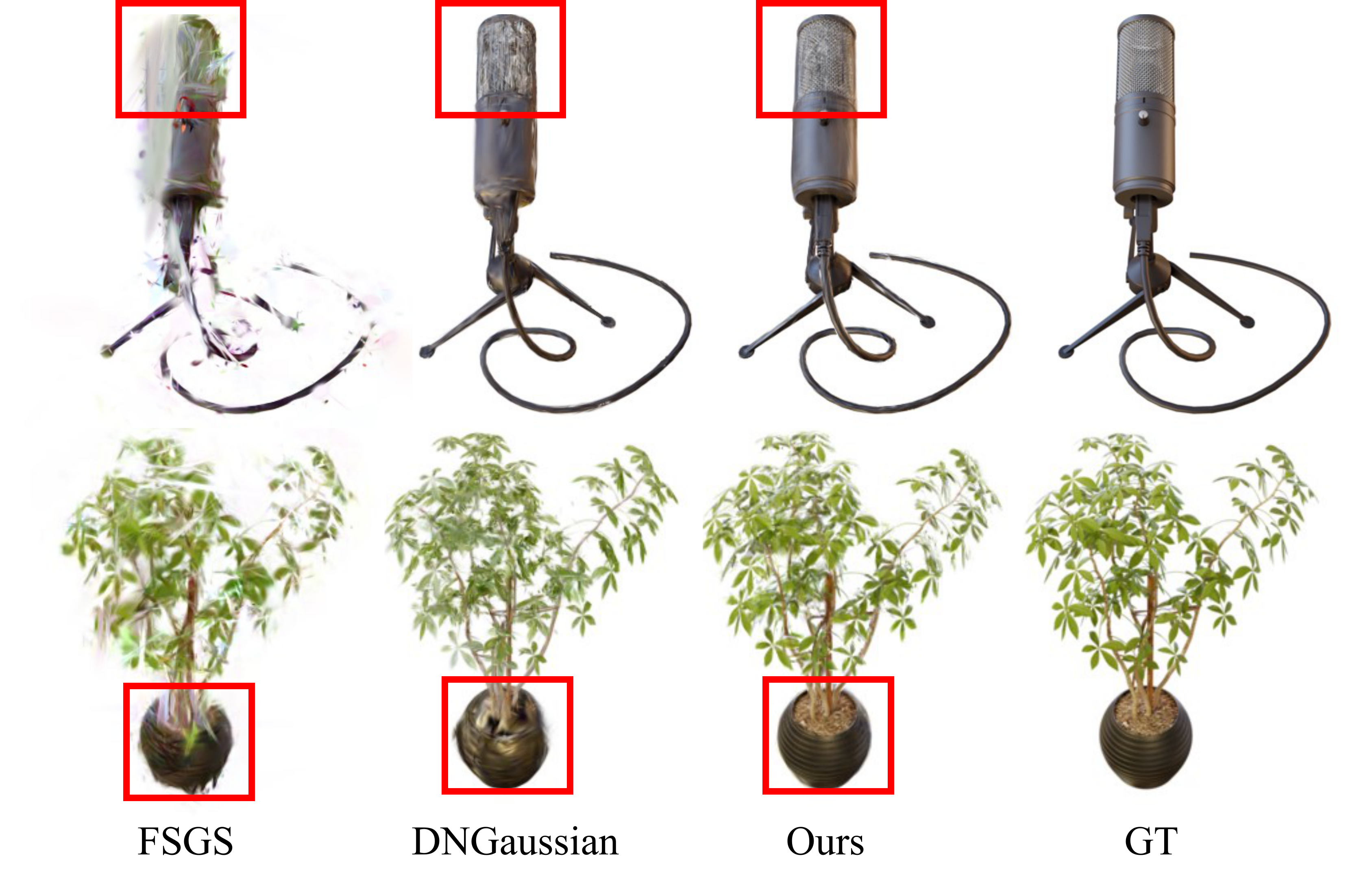}
    \caption{Visual comparison on Blender dataset.}
    \vspace{-0.5cm}
    \label{fig:blender_visual}
\end{wrapfigure}

Figure \ref{fig:llff_visual} presents the visual comparison of novel view synthesis and depth rendering for the \textit{leaves} and \textit{orchids} scenes in the LLFF dataset. From the rendered results of novel views, several state-of-the-art methods are all perceptually acceptable. However, regarding the depth maps, RegNerf \cite{regnerf}, FreeNerf \cite{freenerf}, and FSGS \cite{fsgs} exhibit poor quality, indicating significant errors in their geometric estimation. SparseNeRF \cite{sparsenerf} and DNGaussian \cite{dngaussian} indirectly utilize relative depth information from depth priors to supervise geometry estimation, hence achieving adequate rendering depth. Although FSGS \cite{fsgs} also exploits depth priors, it
attempts to directly employ depth prior information by using depth correlation loss. It does not yield accurate scene geometry, instead, inaccurate depth prior information exacerbates scene geometry. In contrast, our method utilizes self-supervised view consistency loss to obtain more precise geometry, thereby recovering more structural details in novel views.

\paragraph{DTU. } Table \ref{tab:dtu} shows the quantitative results of the DTU dataset under 3, 6, and 9 input views respectively. MuRF \cite{xu2024murf} performs better on the DTU dataset when using 3 views as input. However, generalizable methods require large scale datasets and time-consuming training to learn a prior. Figure \ref{fig:dtu_visual} illustrates the visual comparison of novel view synthesis for three scenes in the DTU dataset. To distinctly discern the differences among various methods, we select the synthesized images far from the training views for comparison in each scene. From the rendered results of novel views, it can be observed that NeRF-based methods produce blurry results, while FSGS \cite{fsgs} and DNGaussian \cite{dngaussian} exhibit numerous artifacts due to insufficient constraints on Gaussians. Our method outperforms previous state-of-the-art methods on most evaluation metrics and achieves better visual quality as well.

\paragraph{Blender. }We evaluate our method on 360-degree scenes using the Blender dataset. Table \ref{tab:blender} presents the quantitative results under 8 input views. Our approach outperforms all baselines in the PSNR score, and achieves comparable SSIM and LPIPS. We could not reproduce the results reported in the FSGS \cite{fsgs} on the Blender dataset and used the own test results in Table \ref{tab:blender}. Figure \ref{fig:blender_visual} illustrates the visual comparison results of several 3DGS-based methods. It can be observed that FSGS \cite{fsgs} exhibits noticeable artifacts in the synthesized images due to insufficient Gaussian constraints, and the rendering results of DNGaussian \cite{dngaussian} also has some distortions. In contrast, our method performs better in terms of detail preservation.

\begin{table}[]
    \caption{Ablation studies on LLFF and DTU dataset with 3 input views.}
    \centering
    \resizebox{0.8\linewidth}{!}{
    \begin{tabular}{cccllllll}
    \hline
    \multirow{2}{*}{Dense Init} & \multirow{2}{*}{$L_{consis}$} & \multirow{2}{*}{Opacity Decay} & \multicolumn{3}{c}{LLFF} & \multicolumn{3}{c}{DTU} \\
     &   &            & PSNR$\uparrow$   & SSIM$\uparrow$   & LPIPS$\downarrow$  & PSNR$\uparrow$   & SSIM$\uparrow$   & LPIPS$\downarrow$ \\ \hline
     &        &       & 15.52  & 0.405  & 0.408  & 10.99  & 0.585  & 0.313 \\
     \checkmark &  &  & 19.30  & 0.651  & 0.238  & 16.36  & 0.778  & 0.181 \\
     & \checkmark  &  & 17.53  & 0.522  & 0.322  & 13.41  & 0.773  & 0.202 \\
     & & \checkmark  & 18.95  & 0.606  & 0.268  & 14.40  & 0.723  & 0.219 \\
     \checkmark & \checkmark &  & 20.48  & 0.715  & 0.218  & \cellcolor{orange}19.08  & \cellcolor{orange}0.832  & \cellcolor{orange}0.131 \\
    & \checkmark & \checkmark  & \cellcolor{yellow}20.82  & \cellcolor{yellow}0.716  & \cellcolor{yellow}0.189  & 15.33  & 0.739  & 0.207 \\
    \checkmark & & \checkmark  & \cellcolor{orange}21.13  & \cellcolor{orange}0.738  & \cellcolor{orange}0.180  & \cellcolor{yellow}17.12  & \cellcolor{yellow}0.812  & \cellcolor{yellow}0.152 \\
    \checkmark & \checkmark & \checkmark & \cellcolor{red}21.44  & \cellcolor{red}0.751  & \cellcolor{red}0.168  & \cellcolor{red}20.71  & \cellcolor{red}0.862  & \cellcolor{red}0.111 \\ \hline
    \end{tabular}
    }
    \label{tab:ablation_llff_dtu}
\end{table}

\begin{wraptable}{r}{0.5\linewidth}
    \vspace{-1.0cm}
    \caption{Ablation studies on Blender dataset with 3 input views.}
    \label{tab:ablation_blender}
    \centering
    \resizebox{\linewidth}{!}{
    \begin{tabular}{ccccc}
    \toprule
        $L_{consis}$ & Opacity Decay  & PSNR $\uparrow$ & SSIM $\uparrow$ & LPIPS $\downarrow$ \\
    \midrule
                        &        & 22.23   & 0.858   & 0.114   \\
            \checkmark &       & \cellcolor{yellow}23.47   & \cellcolor{yellow}0.861   & \cellcolor{yellow}0.112   \\
            & \checkmark       & \cellcolor{orange}23.96   & \cellcolor{orange}0.865   & \cellcolor{orange}0.109   \\
            \checkmark & \checkmark   & \cellcolor{red}24.71   & \cellcolor{red}0.872   & \cellcolor{red}0.101   \\
    \bottomrule
    \end{tabular}}
    \vspace{-0.3cm}
\end{wraptable}

\subsection{Ablation Study}
To verify the effectiveness of the view consistency loss, opacity decay strategy and the dense initialization for Gaussians, we isolate each of these modules separately while keeping the other modules unchanged. We then evaluate the metrics and illustrate the visual results. As shown in Tables \ref{tab:ablation_llff_dtu} and \ref{tab:ablation_blender}, the performance decreases when removing any of the modules we proposed.

\begin{figure}
    \centering
    \includegraphics[width=1.0\linewidth]{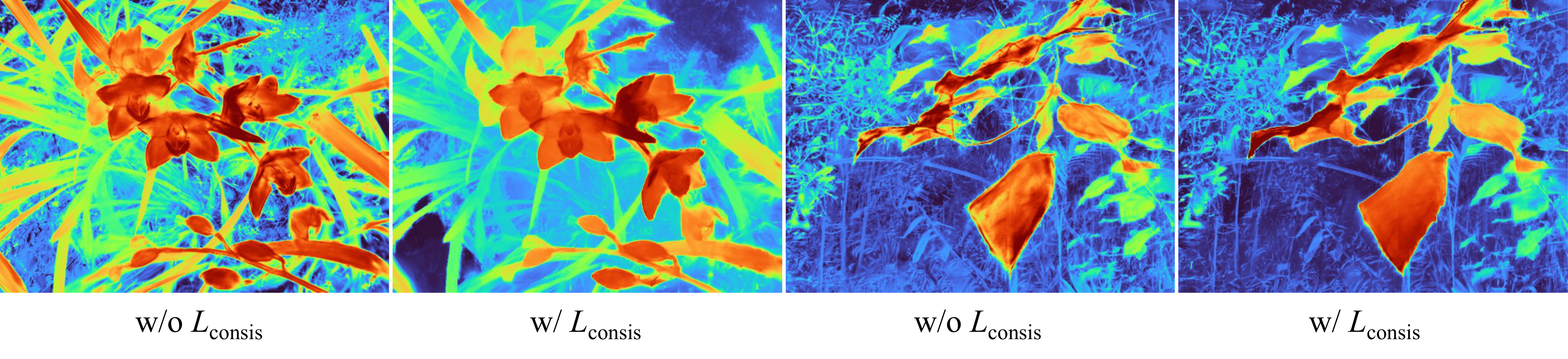}
    \caption{Visual comparison of depth maps before and after using view consistency loss.}
    \label{fig:ablation_view_consis}
\end{figure}

\paragraph{Effectiveness of view consistency loss.} 
To verify the effectiveness of the view consistency loss, we compare the depth maps from novel views. Figure \ref{fig:ablation_view_consis} shows the visual comparison of rendered depth maps. The comparison in the figure shows that it is evident that the view consistency loss significantly improves the estimation of scene geometry.

\begin{figure}[]
    \centering
    \includegraphics[width=0.8\linewidth]{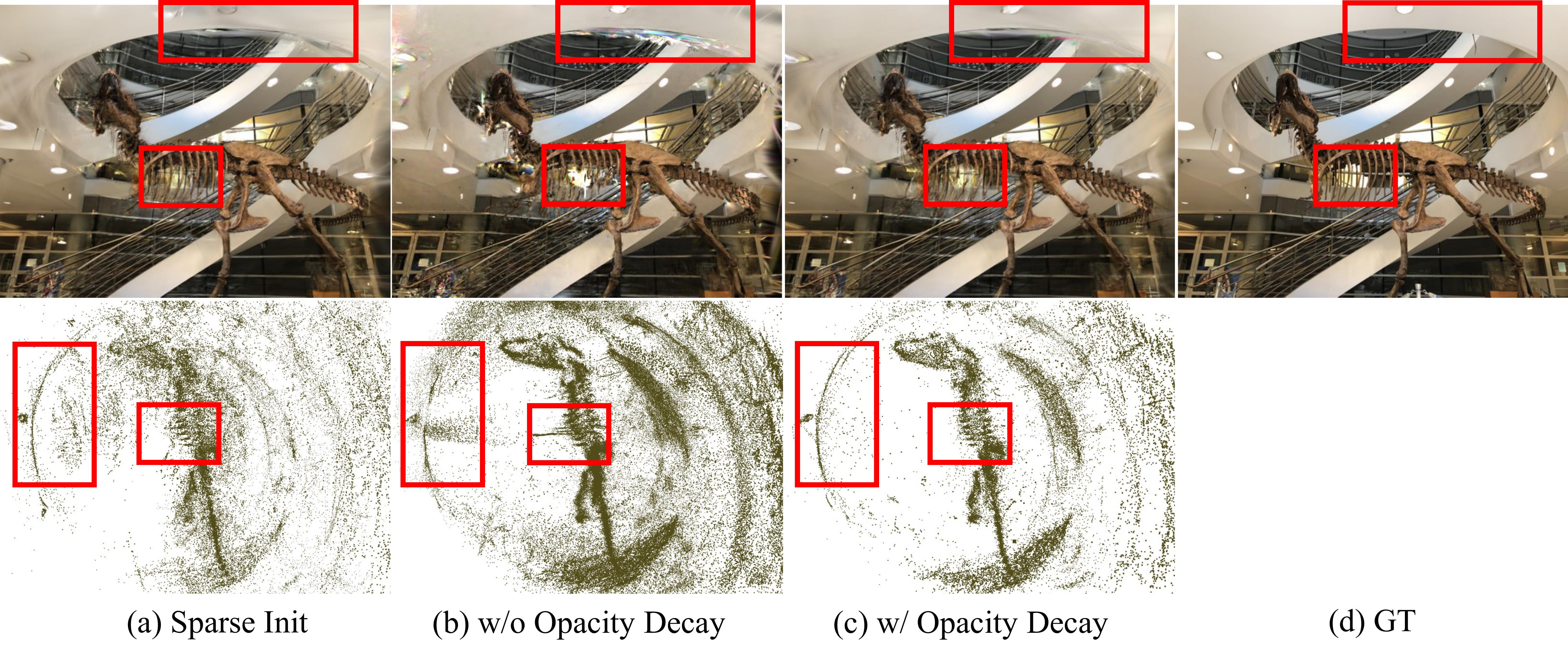}
    \caption{Visual comparison of novel view images and Gaussian point clouds.}
    \label{fig:ablation_opa_and_init}
\end{figure}

\paragraph{Effectiveness of opacity decay. }Figure \ref{fig:ablation_opa_and_init} (b) and (c) show the comparison of novel view synthesis results and the visualization of Gaussian centers for the \textit{trex} scene in LLFF before and after applying the opacity decay. It can be seen that without the opacity decay, Gaussians appear far from the surfaces and there is noise near the surfaces. After applying the opacity decay strategy, the Gaussians far from the scene surfaces are eliminated, and the noisy point clouds are significantly reduced, thereby further improving the rendering quality of novel views.

\paragraph{Effectiveness of dense initialization. }Figure \ref{fig:ablation_opa_and_init} (a) and (c) show the novel view images and the final Gaussian point clouds obtained using different initialization point cloud. It can be seen that without dense point cloud initialization, the spatial positions of the Gaussians become arbitrary in some low-texture regions or areas occluded in the training views due to insufficient constraints, resulting in reduced quality of the novel view images.

\section{Conclusion}
In this paper, we propose a novel method for novel view synthesis from sparse views with 3DGS. We construct a self-supervised multi-view consistency constraint using the rendered and input images, and introduce a Gaussian opacity decay and a dense point cloud initialization strategy. These constraints ensure that the Gaussians are distributed as closely as possible to the scene surfaces and filter out those far from the surfaces. Our approach enables the unstructured Gaussians to accurately represent scene geometry even with sparse input views, resulting in high-quality novel rendering images. Extensive experiments on the LLFF, DTU and Blender datasets demonstrate that our method outperforms existing state-of-the-art sparse view synthesis methods.

\paragraph{Limitation.} 
Since our method utilizes the view consistency constraints for estimating scene depth, some scene areas with low texture may lead to the inaccurate depth estimation (e.g. the white background areas in the DTU dataset), thus failing to constrain the corresponding Gaussians. This results in the white Gaussians potentially occluding the object in the novel views. In contrast, DNGaussian \cite{dngaussian} uses the depth prior estimated by the pre-trained network to constrain the Gaussians, preventing this scenario from happening. The visual comparisons are presented in the Appendix.

\section{Acknowledgement}
This work was supported by National Key R\&D Program of China (2022YFC3800600), the National Natural Science Foundation of China (62272263, 62072268), and in part by Tsinghua-Kuaishou Institute of Future Media Data.

% \section*{References}
\bibliographystyle{plain}
\bibliography{papers}

\begin{thebibliography}{10}

\bibitem{mipnerf}
Jonathan~T Barron, Ben Mildenhall, Matthew Tancik, Peter Hedman, Ricardo Martin-Brualla, and Pratul~P Srinivasan.
\newblock {Mip-NeRF}: A multiscale representation for anti-aliasing neural radiance fields.
\newblock In {\em Proceedings of the IEEE/CVF International Conference on Computer Vision}, pages 5855--5864, 2021.

\bibitem{mipnerf360}
Jonathan~T Barron, Ben Mildenhall, Dor Verbin, Pratul~P Srinivasan, and Peter Hedman.
\newblock {Mip-NeRF 360}: Unbounded anti-aliased neural radiance fields.
\newblock In {\em Proceedings of the IEEE/CVF Conference on Computer Vision and Pattern Recognition}, pages 5470--5479, 2022.

\bibitem{zip-nerf}
Jonathan~T Barron, Ben Mildenhall, Dor Verbin, Pratul~P Srinivasan, and Peter Hedman.
\newblock {Zip-NeRF}: Anti-aliased grid-based neural radiance fields.
\newblock In {\em Proceedings of the IEEE/CVF International Conference on Computer Vision}, pages 19697--19705, 2023.

\bibitem{tensorf}
Anpei Chen, Zexiang Xu, Andreas Geiger, Jingyi Yu, and Hao Su.
\newblock {TensoRF}: Tensorial radiance fields.
\newblock In {\em European Conference on Computer Vision}, pages 333--350. Springer, 2022.

\bibitem{mvsnerf}
Anpei Chen, Zexiang Xu, Fuqiang Zhao, Xiaoshuai Zhang, Fanbo Xiang, Jingyi Yu, and Hao Su.
\newblock {MVSNeRF}: Fast generalizable radiance field reconstruction from multi-view stereo.
\newblock In {\em Proceedings of the IEEE/CVF International Conference on Computer Vision}, pages 14124--14133, 2021.

\bibitem{geoaug}
Di~Chen, Yu~Liu, Lianghua Huang, Bin Wang, and Pan Pan.
\newblock {GeoAug}: Data augmentation for few-shot nerf with geometry constraints.
\newblock In {\em European Conference on Computer Vision}, pages 322--337. Springer, 2022.

\bibitem{chou2023diffusion}
Gene Chou, Yuval Bahat, and Felix Heide.
\newblock {Diffusion-SDF}: Conditional generative modeling of signed distance functions.
\newblock In {\em Proceedings of the IEEE/CVF International Conference on Computer Vision}, pages 2262--2272, 2023.

\bibitem{depthnerf}
Kangle Deng, Andrew Liu, Jun-Yan Zhu, and Deva Ramanan.
\newblock {Depth-supervised NeRF}: Fewer views and faster training for free.
\newblock In {\em Proceedings of the IEEE/CVF Conference on Computer Vision and Pattern Recognition}, pages 12882--12891, 2022.

\bibitem{k-planes}
Sara Fridovich-Keil, Giacomo Meanti, Frederik~Rahb{\ae}k Warburg, Benjamin Recht, and Angjoo Kanazawa.
\newblock {K-Planes}: Explicit radiance fields in space, time, and appearance.
\newblock In {\em Proceedings of the IEEE/CVF Conference on Computer Vision and Pattern Recognition}, pages 12479--12488, 2023.

\bibitem{plenoxels}
Sara Fridovich-Keil, Alex Yu, Matthew Tancik, Qinhong Chen, Benjamin Recht, and Angjoo Kanazawa.
\newblock Plenoxels: Radiance fields without neural networks.
\newblock In {\em Proceedings of the IEEE/CVF Conference on Computer Vision and Pattern Recognition}, pages 5501--5510, 2022.

\bibitem{godard2017binocular}
Cl{\'e}ment Godard, Oisin Mac~Aodha, and Gabriel~J Brostow.
\newblock Unsupervised monocular depth estimation with left-right consistency.
\newblock In {\em Proceedings of the IEEE Conference on Computer Vision and Pattern Recognition}, pages 270--279, 2017.

\bibitem{diggingbinocular}
Cl{\'e}ment Godard, Oisin Mac~Aodha, Michael Firman, and Gabriel~J Brostow.
\newblock Digging into self-supervised monocular depth estimation.
\newblock In {\em Proceedings of the IEEE/CVF International Conference on Computer Vision}, pages 3828--3838, 2019.

\bibitem{guedon2023sugar}
Antoine Gu{\'e}don and Vincent Lepetit.
\newblock {SuGaR}: Surface-aligned gaussian splatting for efficient {3D} mesh reconstruction and high-quality mesh rendering.
\newblock In {\em Proceedings of the IEEE/CVF Conference on Computer Vision and Pattern Recognition}, pages 5354--5363, 2024.

\bibitem{triangulation}
Richard Hartley and Andrew Zisserman.
\newblock {\em Multiple View Geometry in Computer Vision}.
\newblock Cambridge university press, 2003.

\bibitem{tri-mipnerf}
Wenbo Hu, Yuling Wang, Lin Ma, Bangbang Yang, Lin Gao, Xiao Liu, and Yuewen Ma.
\newblock {Tri-MipRF}: {Tri-Mip} representation for efficient anti-aliasing neural radiance fields.
\newblock In {\em Proceedings of the IEEE/CVF International Conference on Computer Vision}, pages 19774--19783, 2023.

\bibitem{huang2023neusurf}
Han Huang, Yulun Wu, Junsheng Zhou, Ge~Gao, Ming Gu, and Yu-Shen Liu.
\newblock {NeuSurf}: On-surface priors for neural surface reconstruction from sparse input views.
\newblock In {\em Proceedings of the AAAI Conference on Artificial Intelligence}, 2024.

\bibitem{dietnerf}
Ajay Jain, Matthew Tancik, and Pieter Abbeel.
\newblock Putting {NeRF on a Diet}: Semantically consistent few-shot view synthesis.
\newblock In {\em Proceedings of the IEEE/CVF International Conference on Computer Vision}, pages 5885--5894, 2021.

\bibitem{dtu}
Rasmus Jensen, Anders Dahl, George Vogiatzis, Engin Tola, and Henrik Aan{\ae}s.
\newblock Large scale multi-view stereopsis evaluation.
\newblock In {\em Proceedings of the IEEE Conference on Computer Vision and Pattern Recognition}, pages 406--413, 2014.

\bibitem{3DGS}
Bernhard Kerbl, Georgios Kopanas, Thomas Leimk{\"u}hler, and George Drettakis.
\newblock {3D} gaussian splatting for real-time radiance field rendering.
\newblock {\em ACM Transactions on Graphics}, 42(4):1--14, 2023.

\bibitem{infonerf}
Mijeong Kim, Seonguk Seo, and Bohyung Han.
\newblock {InfoNeRF}: Ray entropy minimization for few-shot neural volume rendering.
\newblock In {\em Proceedings of the IEEE/CVF Conference on Computer Vision and Pattern Recognition}, pages 12912--12921, 2022.

\bibitem{geconerf}
Min-Seop Kwak, Jiuhn Song, and Seungryong Kim.
\newblock {GeCoNeRF}: Few-shot neural radiance fields via geometric consistency.
\newblock 2023.

\bibitem{dngaussian}
Jiahe Li, Jiawei Zhang, Xiao Bai, Jin Zheng, Xin Ning, Jun Zhou, and Lin Gu.
\newblock {DNGaussian}: Optimizing sparse-view {3D} gaussian radiance fields with global-local depth normalization.
\newblock In {\em Proceedings of the IEEE/CVF Conference on Computer Vision and Pattern Recognition}, pages 20775--20785, 2024.

\bibitem{li2023neaf}
Shujuan Li, Junsheng Zhou, Baorui Ma, Yu-Shen Liu, and Zhizhong Han.
\newblock {NeAF}: Learning neural angle fields for point normal estimation.
\newblock In {\em Proceedings of the AAAI Conference on Artificial Intelligence}, 2023.

\bibitem{li2024LDI}
Shujuan Li, Junsheng Zhou, Baorui Ma, Yu-Shen Liu, and Zhizhong Han.
\newblock Learning continuous implicit field with local distance indicator for arbitrary-scale point cloud upsampling.
\newblock In {\em Proceedings of the AAAI Conference on Artificial Intelligence}, 2024.

\bibitem{ma2023geodream}
Baorui Ma, Haoge Deng, Junsheng Zhou, Yu-Shen Liu, Tiejun Huang, and Xinlong Wang.
\newblock {GeoDream: Disentangling {2D} and geometric priors for high-fidelity and consistent {3D} generation}.
\newblock {\em arXiv preprint arXiv:2311.17971}, 2023.

\bibitem{ma2023towards}
Baorui Ma, Junsheng Zhou, Yu-Shen Liu, and Zhizhong Han.
\newblock Towards better gradient consistency for neural signed distance functions via level set alignment.
\newblock In {\em Proceedings of the IEEE/CVF Conference on Computer Vision and Pattern Recognition}, pages 17724--17734, 2023.

\bibitem{llff}
Ben Mildenhall, Pratul~P Srinivasan, Rodrigo Ortiz-Cayon, Nima~Khademi Kalantari, Ravi Ramamoorthi, Ren Ng, and Abhishek Kar.
\newblock Local light field fusion: Practical view synthesis with prescriptive sampling guidelines.
\newblock {\em ACM Transactions on Graphics}, 38(4):1--14, 2019.

\bibitem{NeRF}
Ben Mildenhall, Pratul~P Srinivasan, Matthew Tancik, Jonathan~T Barron, Ravi Ramamoorthi, and Ren Ng.
\newblock {NeRF}: Representing scenes as neural radiance fields for view synthesis.
\newblock {\em Communications of the ACM}, 65(1):99--106, 2021.

\bibitem{instant-ngp}
Thomas M{\"u}ller, Alex Evans, Christoph Schied, and Alexander Keller.
\newblock Instant neural graphics primitives with a multiresolution hash encoding.
\newblock {\em ACM Transactions on Graphics}, 41(4):1--15, 2022.

\bibitem{regnerf}
Michael Niemeyer, Jonathan~T Barron, Ben Mildenhall, Mehdi~SM Sajjadi, Andreas Geiger, and Noha Radwan.
\newblock {RegNeRF}: Regularizing neural radiance fields for view synthesis from sparse inputs.
\newblock In {\em Proceedings of the IEEE/CVF Conference on Computer Vision and Pattern Recognition}, pages 5480--5490, 2022.

\bibitem{coherentgs}
Avinash Paliwal, Wei Ye, Jinhui Xiong, Dmytro Kotovenko, Rakesh Ranjan, Vikas Chandra, and Nima~Khademi Kalantari.
\newblock {CoherentGS}: Sparse novel view synthesis with coherent {3D} gaussians.
\newblock In {\em European Conference on Computer Vision}, 2024.

\bibitem{transformer4depth}
Ren{\'e} Ranftl, Alexey Bochkovskiy, and Vladlen Koltun.
\newblock Vision transformers for dense prediction.
\newblock In {\em Proceedings of the IEEE/CVF International Conference on Computer Vision}, pages 12179--12188, 2021.

\bibitem{robust_depth}
Ren{\'e} Ranftl, Katrin Lasinger, David Hafner, Konrad Schindler, and Vladlen Koltun.
\newblock Towards robust monocular depth estimation: Mixing datasets for zero-shot cross-dataset transfer.
\newblock {\em IEEE Transactions on Pattern Analysis and Machine Intelligence}, 44(3):1623--1637, 2020.

\bibitem{safadoust2024self}
Sadra Safadoust, Fabio Tosi, Fatma G{\"u}ney, and Matteo Poggi.
\newblock Self-evolving depth-supervised {3D} gaussian splatting from rendered stereo pairs.
\newblock 2024.

\bibitem{sfm}
Johannes~L Schonberger and Jan-Michael Frahm.
\newblock Structure-from-motion revisited.
\newblock In {\em Proceedings of the IEEE conference on Computer Vision and Pattern Recognition}, pages 4104--4113, 2016.

\bibitem{flipnerf}
Seunghyeon Seo, Yeonjin Chang, and Nojun Kwak.
\newblock {FlipNeRF}: Flipped reflection rays for few-shot novel view synthesis.
\newblock In {\em Proceedings of the IEEE/CVF International Conference on Computer Vision}, pages 22883--22893, 2023.

\bibitem{vipnerf}
Nagabhushan Somraj and Rajiv Soundararajan.
\newblock {Vip-NeRF}: Visibility prior for sparse input neural radiance fields.
\newblock In {\em ACM SIGGRAPH 2023 Conference Proceedings}, pages 1--11, 2023.

\bibitem{darf}
Jiuhn Song, Seonghoon Park, Honggyu An, Seokju Cho, Min-Seop Kwak, Sungjin Cho, and Seungryong Kim.
\newblock D{\"a}{RF}: Boosting radiance fields from sparse input views with monocular depth adaptation.
\newblock {\em Advances in Neural Information Processing Systems}, 36, 2024.

\bibitem{dvgo}
Cheng Sun, Min Sun, and Hwann-Tzong Chen.
\newblock Direct voxel grid optimization: Super-fast convergence for radiance fields reconstruction.
\newblock In {\em Proceedings of the IEEE/CVF Conference on Computer Vision and Pattern Recognition}, pages 5459--5469, 2022.

\bibitem{sun2023vgos}
Jiakai Sun, Zhanjie Zhang, Jiafu Chen, Guangyuan Li, Boyan Ji, Lei Zhao, and Wei Xing.
\newblock {VGOS}: Voxel grid optimization for view synthesis from sparse inputs.
\newblock In {\em Proceedings of the Thirty-Second International Joint Conference on Artificial Intelligence}, pages 1414--1422, 2023.

\bibitem{loftr}
Jiaming Sun, Zehong Shen, Yuang Wang, Hujun Bao, and Xiaowei Zhou.
\newblock {LoFTR}: Detector-free local feature matching with transformers.
\newblock In {\em Proceedings of the IEEE/CVF Conference on Computer Vision and Pattern Recognition}, pages 8922--8931, 2021.

\bibitem{pdcnet_plus}
Prune Truong, Martin Danelljan, Radu Timofte, and Luc Van~Gool.
\newblock {PDC-Net+}: Enhanced probabilistic dense correspondence network.
\newblock {\em IEEE Transactions on Pattern Analysis and Machine Intelligence}, 2023.

\bibitem{sparf}
Prune Truong, Marie-Julie Rakotosaona, Fabian Manhardt, and Federico Tombari.
\newblock {SPARF}: Neural radiance fields from sparse and noisy poses.
\newblock In {\em Proceedings of the IEEE/CVF Conference on Computer Vision and Pattern Recognition}, pages 4190--4200, 2023.

\bibitem{sparsenerf}
Guangcong Wang, Zhaoxi Chen, Chen~Change Loy, and Ziwei Liu.
\newblock {SparseNeRF}: Distilling depth ranking for few-shot novel view synthesis.
\newblock In {\em Proceedings of the IEEE/CVF International Conference on Computer Vision}, pages 9065--9076, 2023.

\bibitem{wang2021neus}
Peng Wang, Lingjie Liu, Yuan Liu, Christian Theobalt, Taku Komura, and Wenping Wang.
\newblock {NeuS}: Learning neural implicit surfaces by volume rendering for multi-view reconstruction.
\newblock {\em Advances in Neural Information Processing Systems}, 34:27171--27183, 2021.

\bibitem{ssim}
Zhou Wang, Alan~C Bovik, Hamid~R Sheikh, and Eero~P Simoncelli.
\newblock Image quality assessment: from error visibility to structural similarity.
\newblock {\em IEEE Transactions on Image Processing}, 13(4):600--612, 2004.

\bibitem{wen20223d}
Xin Wen, Junsheng Zhou, Yu-Shen Liu, Hua Su, Zhen Dong, and Zhizhong Han.
\newblock {3D} shape reconstruction from {2D} images with disentangled attribute flow.
\newblock In {\em Proceedings of the IEEE/CVF Conference on Computer Vision and Pattern Recognition}, pages 3803--3813, 2022.

\bibitem{wolf2024gs2mesh}
Yaniv Wolf, Amit Bracha, and Ron Kimmel.
\newblock {GS2Mesh}: Surface reconstruction from gaussian splatting via novel stereo views.
\newblock In {\em European Conference on Computer Vision}, 2024.

\bibitem{reconfusion}
Rundi Wu, Ben Mildenhall, Philipp Henzler, Keunhong Park, Ruiqi Gao, Daniel Watson, Pratul~P Srinivasan, Dor Verbin, Jonathan~T Barron, Ben Poole, et~al.
\newblock {ReconFusion}: {3D} reconstruction with diffusion priors.
\newblock In {\em Proceedings of the IEEE/CVF Conference on Computer Vision and Pattern Recognition}, pages 21551--21561, 2024.

\bibitem{diffusionerf}
Jamie Wynn and Daniyar Turmukhambetov.
\newblock {DiffusioNeRF}: Regularizing neural radiance fields with denoising diffusion models.
\newblock In {\em Proceedings of the IEEE/CVF Conference on Computer Vision and Pattern Recognition}, pages 4180--4189, 2023.

\bibitem{sparsegs}
Haolin Xiong, Sairisheek Muttukuru, Rishi Upadhyay, Pradyumna Chari, and Achuta Kadambi.
\newblock {SparseGS}: Real-time 360° sparse view synthesis using gaussian splatting.
\newblock {\em arXiv preprint arXiv:2312.00206}, 2023.

\bibitem{xu2024murf}
Haofei Xu, Anpei Chen, Yuedong Chen, Christos Sakaridis, Yulun Zhang, Marc Pollefeys, Andreas Geiger, and Fisher Yu.
\newblock {MuRF}: Multi-baseline radiance fields.
\newblock In {\em Proceedings of the IEEE/CVF Conference on Computer Vision and Pattern Recognition}, pages 20041--20050, 2024.

\bibitem{yan2023implicit}
Siming Yan, Zhenpei Yang, Haoxiang Li, Chen Song, Li~Guan, Hao Kang, Gang Hua, and Qixing Huang.
\newblock Implicit autoencoder for point-cloud self-supervised representation learning.
\newblock In {\em Proceedings of the IEEE/CVF International Conference on Computer Vision}, pages 14530--14542, 2023.

\bibitem{freenerf}
Jiawei Yang, Marco Pavone, and Yue Wang.
\newblock {FreeNeRF}: Improving few-shot neural rendering with free frequency regularization.
\newblock In {\em Proceedings of the IEEE/CVF Conference on Computer Vision and Pattern Recognition}, pages 8254--8263, 2023.

\bibitem{depth_anything}
Lihe Yang, Bingyi Kang, Zilong Huang, Xiaogang Xu, Jiashi Feng, and Hengshuang Zhao.
\newblock {Depth Anything}: Unleashing the power of large-scale unlabeled data.
\newblock pages 10371--10381, 2024.

\bibitem{yariv2021volume}
Lior Yariv, Jiatao Gu, Yoni Kasten, and Yaron Lipman.
\newblock Volume rendering of neural implicit surfaces.
\newblock In {\em Advances in Neural Information Processing Systems}, 2021.

\bibitem{pixelnerf}
Alex Yu, Vickie Ye, Matthew Tancik, and Angjoo Kanazawa.
\newblock {pixelNeRF}: Neural radiance fields from one or few images.
\newblock In {\em Proceedings of the IEEE/CVF Conference on Computer Vision and Pattern Recognition}, pages 4578--4587, 2021.

\bibitem{zhang2023lite-mono}
Ning Zhang, Francesco Nex, George Vosselman, and Norman Kerle.
\newblock {Lite-Mono}: A lightweight cnn and transformer architecture for self-supervised monocular depth estimation.
\newblock In {\em Proceedings of the IEEE/CVF Conference on Computer Vision and Pattern Recognition}, pages 18537--18546, 2023.

\bibitem{lpips}
Richard Zhang, Phillip Isola, Alexei~A Efros, Eli Shechtman, and Oliver Wang.
\newblock The unreasonable effectiveness of deep features as a perceptual metric.
\newblock In {\em Proceedings of the IEEE Conference on Computer Vision and Pattern Recognition}, pages 586--595, 2018.

\bibitem{zhang2024gspull}
Wenyuan Zhang, Yu-Shen Liu, and Zhizhong Han.
\newblock Neural signed distance function inference through splatting {3D} gaussians pulled on zero-level set.
\newblock In {\em Advances in Neural Information Processing Systems}, 2024.

\bibitem{zhou2024deep}
Junsheng Zhou, Yu-Shen Liu, and Zhizhong Han.
\newblock Zero-shot scene reconstruction from single images with deep prior assembly.
\newblock In {\em Advances in Neural Information Processing Systems}, 2024.

\bibitem{zhou2024cap}
Junsheng Zhou, Baorui Ma, Shujuan Li, Yu-Shen Liu, Yi~Fang, and Zhizhong Han.
\newblock {CAP-UDF}: Learning unsigned distance functions progressively from raw point clouds with consistency-aware field optimization.
\newblock {\em IEEE Transactions on Pattern Analysis and Machine Intelligence}, 2024.

\bibitem{zhou2023levelset}
Junsheng Zhou, Baorui Ma, Shujuan Li, Yu-Shen Liu, and Zhizhong Han.
\newblock Learning a more continuous zero level set in unsigned distance fields through level set projection.
\newblock In {\em Proceedings of the IEEE/CVF International Conference on Computer Vision}, 2023.

\bibitem{zhou2024fast}
Junsheng Zhou, Baorui Ma, and Yu-Shen Liu.
\newblock Fast learning of signed distance functions from noisy point clouds via noise to noise mapping.
\newblock {\em IEEE Transactions on Pattern Analysis and Machine Intelligence}, 2024.

\bibitem{Zhou2022CAP-UDF}
Junsheng Zhou, Baorui Ma, Liu Yu-Shen, Fang Yi, and Han Zhizhong.
\newblock Learning consistency-aware unsigned distance functions progressively from raw point clouds.
\newblock In {\em Advances in Neural Information Processing Systems}, 2022.

\bibitem{Zhou2023VP2P}
Junsheng Zhou, Baorui Ma, Wenyuan Zhang, Yi~Fang, Yu-Shen Liu, and Zhizhong Han.
\newblock Differentiable registration of images and lidar point clouds with voxelpoint-to-pixel matching.
\newblock In {\em Advances in Neural Information Processing Systems}, 2023.

\bibitem{zhou2023uni3d}
Junsheng Zhou, Jinsheng Wang, Baorui Ma, Yu-Shen Liu, Tiejun Huang, and Xinlong Wang.
\newblock {Uni3D: Exploring Unified 3D Representation at Scale}.
\newblock In {\em International Conference on Learning Representations}, 2024.

\bibitem{zhou20223d}
Junsheng Zhou, Xin Wen, Baorui Ma, Yu-Shen Liu, Yue Gao, Yi~Fang, and Zhizhong Han.
\newblock {3D-OAE}: Occlusion auto-encoders for self-supervised learning on point clouds.
\newblock {\em IEEE International Conference on Robotics and Automation}, 2024.

\bibitem{zhou2024diffgs}
Junsheng Zhou, Weiqi Zhang, and Yu-Shen Liu.
\newblock {DiffGS}: Functional gaussian splatting diffusion.
\newblock In {\em Advances in Neural Information Processing Systems}, 2024.

\bibitem{udiff}
Junsheng Zhou, Weiqi Zhang, Baorui Ma, Kanle Shi, Yu-Shen Liu, and Zhizhong Han.
\newblock {UDiFF}: Generating conditional unsigned distance fields with optimal wavelet diffusion.
\newblock In {\em Proceedings of the IEEE/CVF Conference on Computer Vision and Pattern Recognition}, 2024.

\bibitem{fsgs}
Zehao Zhu, Zhiwen Fan, Yifan Jiang, and Zhangyang Wang.
\newblock {FSGS}: Real-time few-shot view synthesis using gaussian splatting.
\newblock In {\em European Conference on Computer Vision}, pages 145--163. Springer, 2024.

\end{thebibliography}

%%%%%%%%%%%%%%%%%%%%%%%%%%%%%%%%%%%%%%%%%%%%%%%%%%%%%%%%%%%%
\newpage
\setcounter{table}{0}
\setcounter{figure}{0}
\renewcommand\thetable{\Alph{table}}
\renewcommand\thefigure{\Alph{figure}}
\appendix

\section{More Visualizations}
\subsection{Additional Results on LLFF}
\begin{figure}
    \centering
    \includegraphics[width=1.0\linewidth]{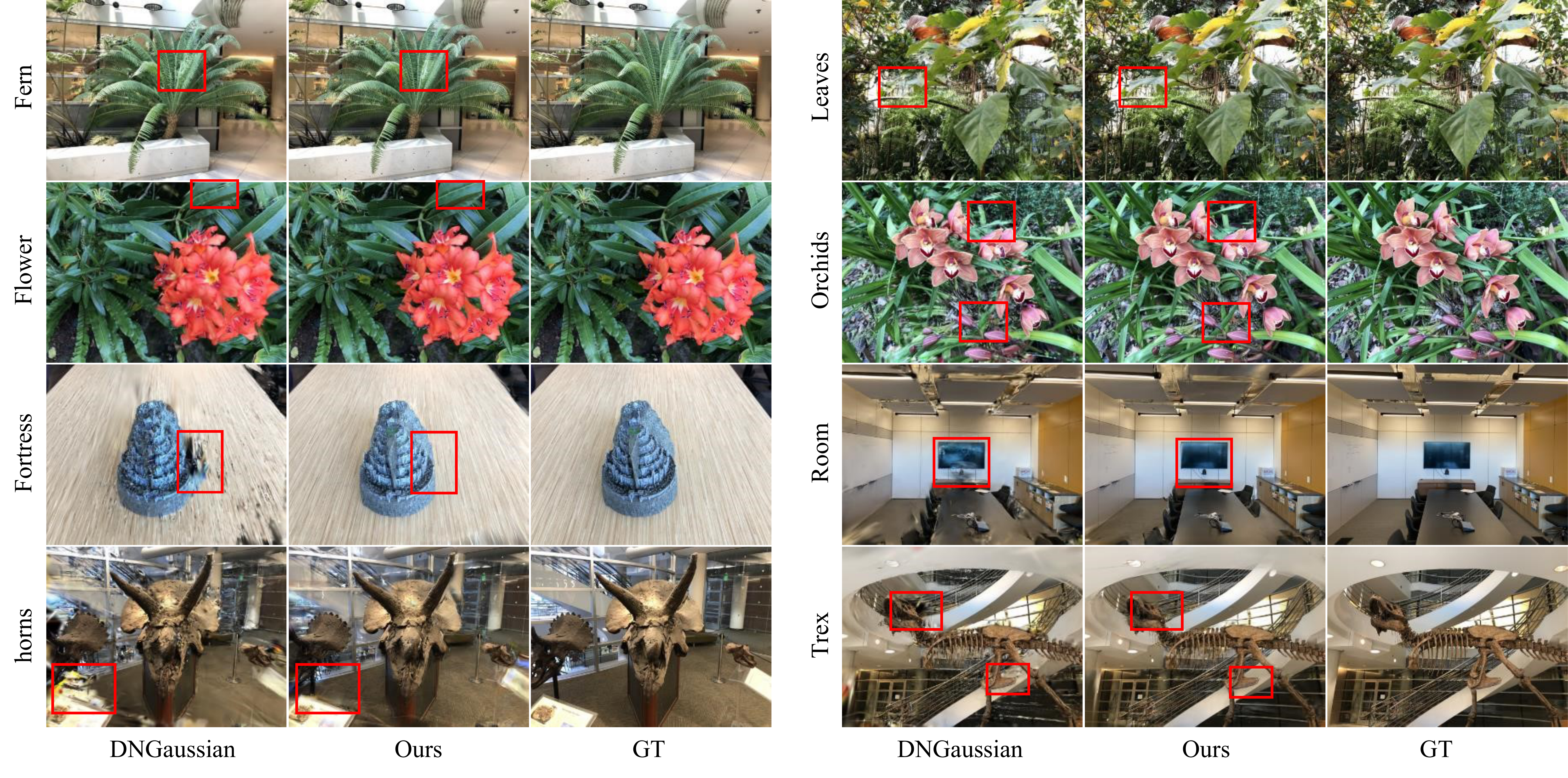}
    \caption{Visual comparison on LLFF dataset with 3 input views.}
    \label{fig:add_llff_visual}
\end{figure}

Figure \ref{fig:add_llff_visual} shows the visualization comparison of our method and DNGaussian \cite{dngaussian} across all scenes in the LLFF \cite{llff} dataset. Each scene is trained using the same 3 input views. Although both methods achieve perceptually acceptable results, the novel view rendering results of our method are closer to the Ground Truth.

\begin{figure}
    \centering
    \includegraphics[width=1.0\linewidth]{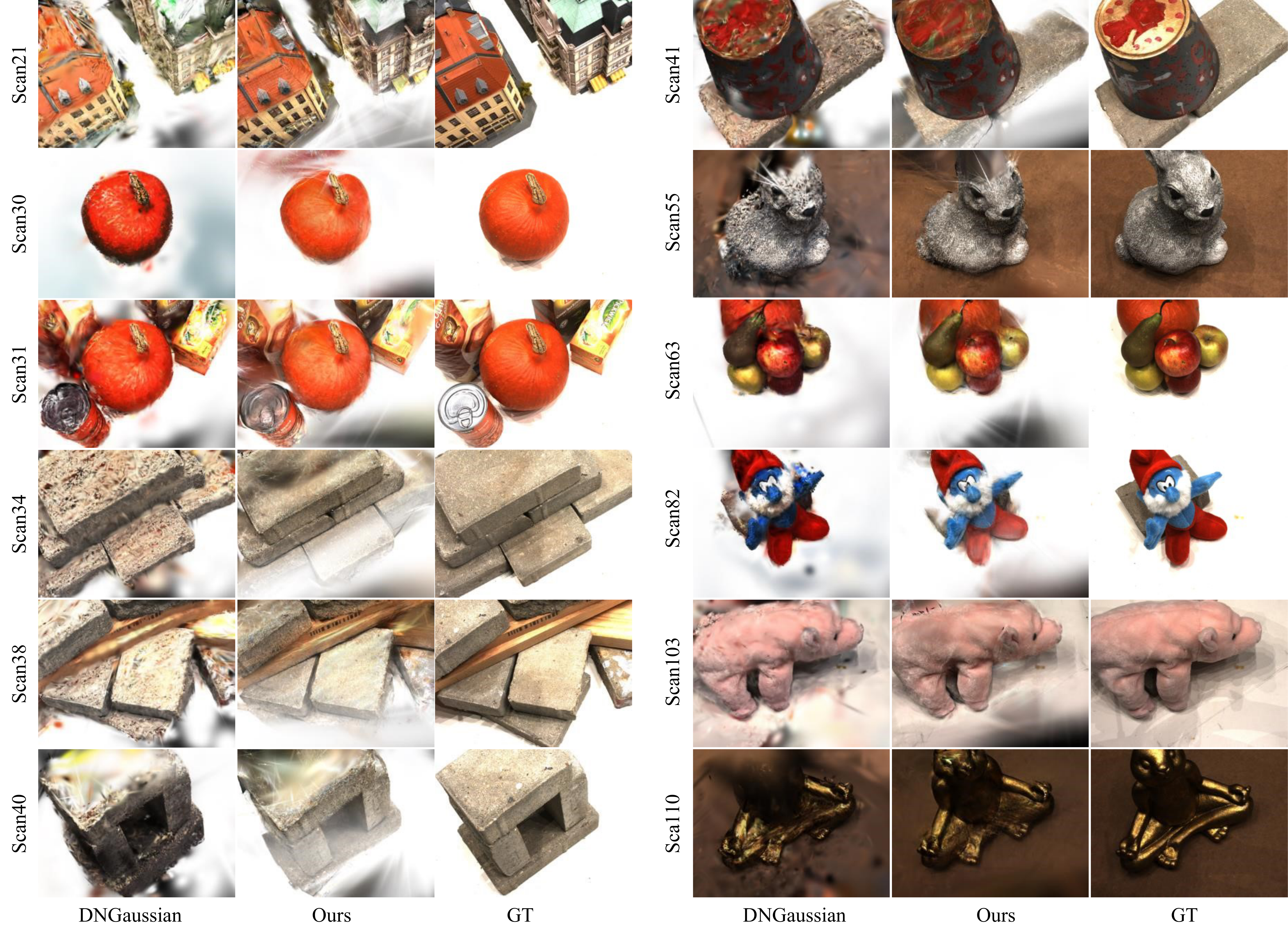}
    \caption{Visual comparison on DTU dataset with 3 input views.}
    \label{fig:add_dtu_visual}
\end{figure}

\subsection{Additional Results on DTU}
Figure \ref{fig:add_dtu_visual} shows an additional visual comparison between our method and DNGaussian \cite{dngaussian} on the DTU \cite{dtu} dataset. Each scene is trained using the same 3 input views. To clearly distinguish the differences between the two methods, we select a novel unseen view that is far from the training views for comparison. As analyzed in the limitations section of the main text, our method cannot effectively constrain the white background in training views, resulting in objects being occluded by white Gaussians in some scenes. Even so, the rendering quality of the novel views from our method is still better than the results from DNGaussian.

\section{More Experiments}
\subsection{Entropy regularization instead of Opacity Decay}
\begin{figure}
    \centering
    \includegraphics[width=0.8\linewidth]{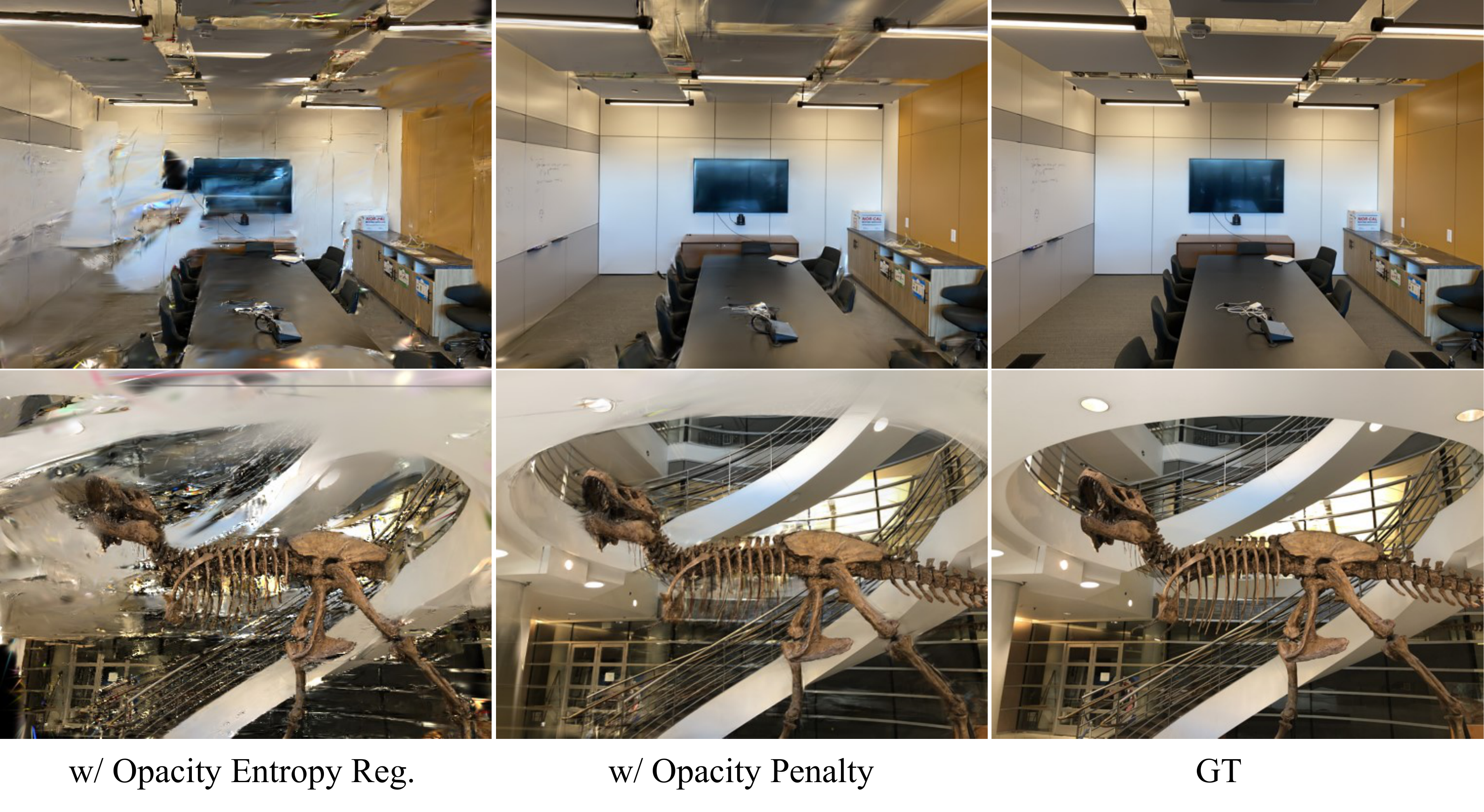}
    \caption{Visual comparison of using different regularization components for opacity.}
    \label{fig:diff_opacity_reg}
\end{figure}

The Sugar \cite{guedon2023sugar} method applies entropy regularization to the opacity of Gaussians to bring Gaussians closer to the surface, similar to our proposed opacity decay strategy. Therefore, we also evaluate the performance of opacity entropy regularization on the LLFF dataset. Following Sugar \cite{guedon2023sugar}, the opacity threshold for Gaussians pruning is set to 0.5. Table \ref{tab:diff_opacity_reg} shows the metrics with opacity entropy regularization. It can be seen that our opacity decay strategy has advantages. Figure \ref{fig:diff_opacity_reg} illustrates the novel view synthesis using different opacity regularization strategies. It can be seen that entropy regularization leads to noticeable overfitting with sparse view inputs, resulting in lower quality novel view rendering.

\begin{table}[]
    \centering
    \begin{tabular}{lccc}
    \toprule
         Components  &  PSNR$\uparrow$ & SSIM$\uparrow$ & LPIPS$\downarrow$ \\
    \midrule
         Opacity Entropy Reg. &  15.75   & 0.480     & 0.361  \\
         Opacity Decay &  \cellcolor{red}21.44  & \cellcolor{red}0.751  & \cellcolor{red}0.168  \\
    \bottomrule
    \end{tabular}
    \caption{Quantitative comparison of using different regularization components for opacity.}
    \label{tab:diff_opacity_reg}
\end{table}

\subsection{Training the DTU with Mask}
\begin{figure}
    \centering
    \includegraphics[width=1.0\linewidth]{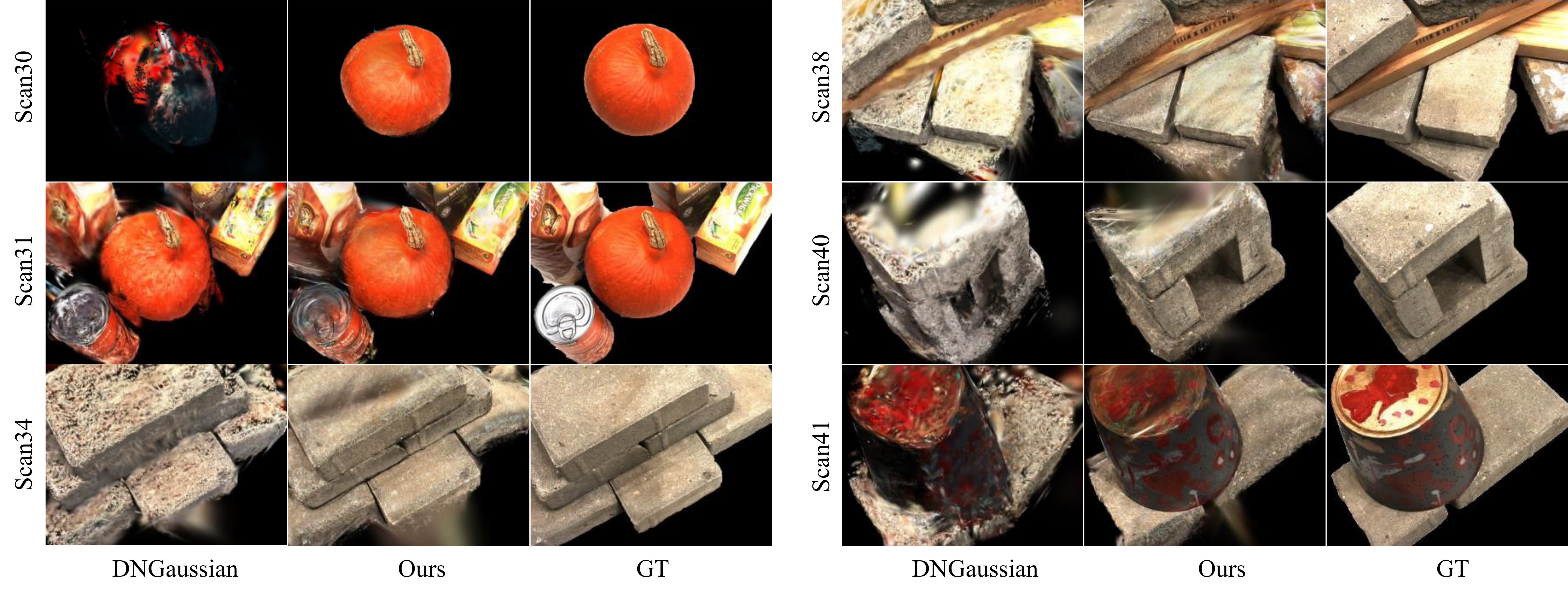}
    \caption{Visual  comparison of using  background masks for input views on DTU dataset.}
    \label{fig:dtu_mask}
\end{figure}

To eliminate the undesirable effects of the solid color background on novel view rendering results in the DTU dataset, we perform additional experiments by using masks to filter out the background in the training views. The evaluation results are shown in Table \ref{tab:dtu_mask}. Figure \ref{fig:dtu_mask} illustrates the visual comparison between our method and DNGaussian \cite{dngaussian} when using masks in the training views. It can be seen that without the solid color background, our method shows a significant improvement, while the performance of DNGaussian decreases.
\begin{table}[]
    \centering
    \begin{tabular}{lccc}
    \toprule
         Methods & PSNR$\uparrow$ & SSIM$\uparrow$ & LPIPS$\downarrow$ \\
    \midrule
         DNGaussian \cite{dngaussian} & 18.91 & 0.790 & 0.176 \\
         DNGaussian$_{mask}$ \cite{dngaussian} & 14.94 & 0.699 & 0.237 \\
         Ours & 20.71 & 0.862 & 0.111\\
         Ours$_{mask}$ & \cellcolor{red}22.03 & \cellcolor{red}0.875 & \cellcolor{red}0.098\\
    \bottomrule
    \end{tabular}
    \caption{Quantitative comparison of using  background masks for input views on DTU dataset.}
    \label{tab:dtu_mask}
\end{table}

\section{Ablation studies for hyperparameters}
\subsection{Ablation studies for hyperparameter $\lambda$}
We perform ablation studies on the LLFF \cite{llff} and DTU \cite{dtu} datasets with the value of $\lambda$ ranging from 0.96 to 1.0, and the results are shown in Table \ref{tab:ablation_lambda}. We find that the best performance is achieved when the value of $\lambda$ is 0.995 on both datasets. Therefore we set $\lambda$ to 0.995 for all datasets.
\begin{table}[h]
    \caption{Ablation studies for $\lambda$ on LLFF dataset with 3 input views.}
    \centering
    \begin{tabular}{lllllllllll}
    \hline
    $\lambda$   &       & 0.960 & 0.970 & 0.975 & 0.980 & 0.985 & 0.990 & 0.995 & 0.998 & 1.0   \\ \hline
    \multirow{3}{*}{LLFF} & PSNR$\uparrow$  & 18.47 & 20.72 & 20.90 & 21.14 & \cellcolor{yellow}21.39 & \cellcolor{orange}21.42 & \cellcolor{red}21.44 & 21.27 & 20.21 \\
    & SSIM$\uparrow$  & 0.629 & 0.711 & 0.712 & 0.730 & 0.742 & \cellcolor{orange}0.748 & \cellcolor{red}0.751 & \cellcolor{yellow}0.745 & 0.709 \\
    & LPIPS$\downarrow$ & 0.344 & 0.237 & 0.223 & 0.202 & 0.184 & \cellcolor{yellow}0.172 & \cellcolor{red}0.168 & \cellcolor{orange}0.171 & 0.201 \\ \hline
    \multirow{3}{*}{DTU}   & PSNR$\uparrow$  & 17.06 & 18.11 & 19.60 & 19.63 & 19.77 & \cellcolor{yellow}20.49 & \cellcolor{red}20.71 & \cellcolor{orange}20.67 & 19.08 \\
    & SSIM$\uparrow$  & 0.785 & 0.813 & 0.846 & 0.848 & 0.853 & \cellcolor{yellow}0.861 & \cellcolor{red}0.862 & \cellcolor{red}0.862 & 0.832 \\
    & LPIPS$\downarrow$ & 0.223 & 0.196 & 0.158 & 0.152 & 0.139 & \cellcolor{yellow}0.121 & \cellcolor{orange}0.111 & \cellcolor{red}0.109 & 0.131 \\ \hline
    \end{tabular}
    \label{tab:ablation_lambda}
\end{table}

\subsection{Ablation studies for hyperparameter $d_{max}$}
We perform ablation studies for $d_{max}$ on the LLFF\cite{llff} and DTU \cite{dtu} datasets. The
value of $d_{max}$ gradually increases from 0.1 to 0.8, we obtain the results as shown in Table \ref{tab:ablation_dmax}.
We find that there is almost no performance increase on both the LLFF and DTU datasets when $d_{max}$ is greater than 0.4, therefore we set the $d_{max}$ to 0.4 for all datasets.
\begin{table}[h]
    \caption{Ablation studies for $d_{max}$ on LLFF and DTU dataset with 3 input views}
    \centering
    \begin{tabular}{lllllll}
    \hline
    \multirow{2}{*}{$d_{max}$} & \multicolumn{3}{c}{LLFF} & \multicolumn{3}{c}{DTU} \\
        & PSNR$\uparrow$   & SSIM$\uparrow$   & LPIPS$\downarrow$  & PSNR$\uparrow$   & SSIM$\uparrow$   & LPIPS$\downarrow$ \\
    \hline
    0.1                   & 21.32  & 0.742  & 0.176  & 19.47  & 0.844  & 0.127 \\
    0.2                   & 21.41  & 0.746  & 0.172  & 19.99  & 0.852  & 0.120 \\
    0.4                   & 21.44  & 0.751  & 0.168  & 20.71  & 0.862  & 0.111 \\
    0.6                   & 21.43  & 0.748  & 0.168  & 20.65  & 0.863  & 0.111 \\
    0.8                   & 21.44  & 0.750  & 0.169  & 20.66  & 0.864  & 0.110 \\
    \hline
    \end{tabular}
    \label{tab:ablation_dmax}
\end{table}

\section{The impact of initialization on overall performance}
We conduct experiments on the LLFF \cite{llff} and DTU \cite{dtu} datasets with several different initialization, including random initialization, sparse initialization, and using the LoFTR \cite{loftr} matching network to construct initial point clouds. The results are shown in Table \ref{tab:diff_init}.

\begin{table}[h]
    \caption{Impact of different initializations on performance.}
    \centering
    \begin{tabular}{lllllll}
    \hline
    \multirow{2}{*}{Init type} & \multicolumn{3}{c}{LLFF} & \multicolumn{3}{c}{DTU} \\
            & PSNR$\uparrow$   & SSIM$\uparrow$   & LPIPS$\downarrow$  & PSNR$\uparrow$   & SSIM$\uparrow$   & LPIPS$\downarrow$ \\
    \hline
    random init                & 17.69  & 0.548  & 0.302  & 16.90  & 0.769  & 0.179 \\
    sparse init                & 20.82  & 0.716  & 0.189  & 15.33  & 0.739  & 0.207 \\
    points by LoFTR \cite{loftr}            & 20.33  & 0.686  & 0.211  & 18.97  & 0.834  & 0.132 \\
    points by PDCNet+ \cite{pdcnet_plus}          & \textbf{21.44}  & \textbf{0.751}  & \textbf{0.168}  & \textbf{20.71}  & \textbf{0.862}  & \textbf{0.111} \\
    \hline
    \end{tabular}
    \label{tab:diff_init}
\end{table}

PDCNet+ \cite{pdcnet_plus} can obtain more matching points than LoFTR \cite{loftr}, so the initialized point cloud obtained by PDCNet+ results in better performance. For the LLFF \cite{llff} dataset, the input views have large overlap, SfM methods can generate point clouds with better quality than LoFTR. Therefore, using sparse point clouds generated by SfM as initialization yields better performance than LoFTR. However, for the DTU \cite{dtu} dataset, where there is small overlap among input views, SfM-generated point clouds are too sparse. Thus, the performance is even worse than random initialization.

\section{The impact of different source views on overall performance}
\begin{figure}
    \centering
    \includegraphics[width=1.0\linewidth]{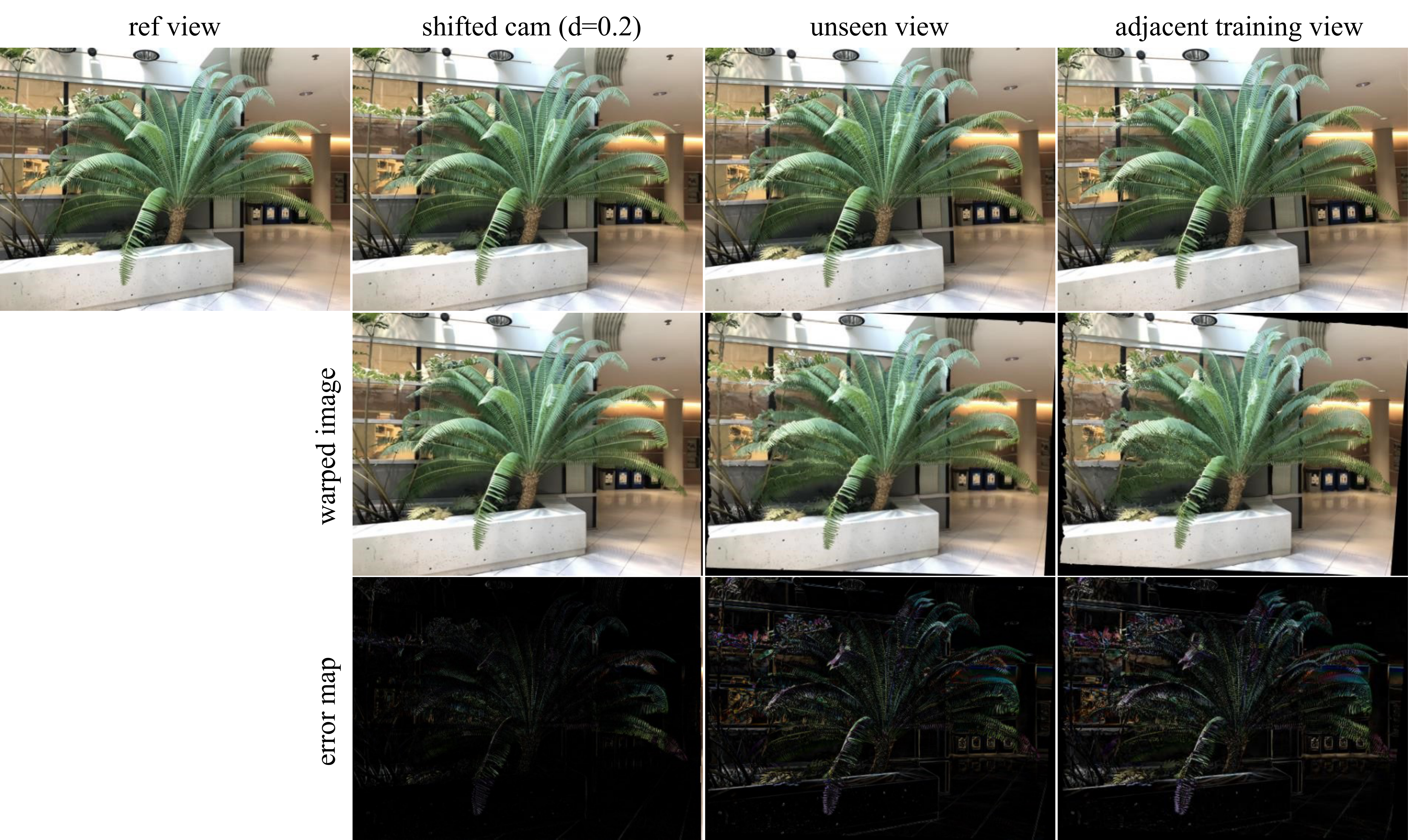}
    \caption{The warped images and error maps when using different source views.}
    \label{fig:warp_compare}
\end{figure}

Figure \ref{fig:warp_compare} shows the warped images and error maps when using different source views. It can be observed that when the source view is rotated relative to the reference view or is far away from the reference view, the warped image suffers from significant distortions due to depth errors and occlusions. This results in a large error compared to the GT image.

Table \ref{tab:warp_compare} shows the quantitative comparison using different source images on the LLFF and DTU datasets. The binocular vision-based view consistency method achieves the best performance.
\begin{table}[h]
    \caption{Quantitative comparison using different source images on the LLFF and the DTU dataset.}
    \centering
    \begin{tabular}{lcccccc}
    \hline
    \multirow{2}{*}{source view type} & \multicolumn{3}{c}{LLFF dataset} & \multicolumn{3}{c}{DTU dataset} \\
    & PSNR$\uparrow$      & SSIM$\uparrow$      & LPIPS$\downarrow$    & PSNR$\uparrow$     & SSIM$\uparrow$     & LPIPS$\downarrow$     \\
    \hline
    adjacent training view         & 20.61     & 0.711     & 0.204    & 19.55    & 0.833 & 0.137    \\
    unseen view                    & 20.88     & 0.728     & 0.185    & 19.96    & 0.850 & 0.124    \\
    shifted cam                    & \textbf{21.44}     & \textbf{0.751}     & \textbf{0.168}    & \textbf{20.71}    &     \textbf{0.862} & \textbf{0.111}    \\
    \hline
    \end{tabular}
    \label{tab:warp_compare}
\end{table}

\section{Implementations Details}
Since our method uses an opacity decay strategy, we do not use the opacity reset strategy from the original 3DGS, and we also remove the scale threshold for Gaussians pruning from the original 3DGS.

We train our model on an RTX 3090 GPU, running 30,000 iterations per scene on the LLFF and DTU datasets, which takes approximately 10 minutes each. For the Blender dataset, we train for 7,000 iterations per scene, which takes approximately 3 minutes each. The storage space required for the Gaussian point clouds is about one-third of that required by the original 3DGS.

\section{Dataset Details}
\paragraph{LLFF.} The LLFF \cite{llff} is a forward-facing dataset with 8 scenes. Following previous works \cite{regnerf, freenerf}, we take every 8th image as the novel views for evaluation. The input views are evenly sampled from the remaining views. During training and evaluation, the images are downsampled by a factor of 8, resulting in a resolution of 378×504 for each image.

\paragraph{DTU.} The DTU \cite{dtu} dataset consists of 124 scenes. We follow previous works \cite{regnerf, freenerf} to train and evaluate our method on 15 test scenes. The test scene IDs are: 8, 21, 30, 31, 34, 38, 40, 41, 45, 55, 63, 82, 103, 110, and 114. In each scene, we use images with the following IDs as input views: 25, 22, 28, 40, 44, 48, 0, 8, 13. The first 3 and 6 image IDs correspond to the input views in 3-view and 6-view settings respectively. We use 25 images as novel views for evaluation, with the following IDs: 1, 2, 9, 10, 11, 12, 14, 15, 23, 24, 26, 27, 29, 30, 31, 32, 33, 34, 35, 41, 42, 43, 45, 46, 47. During training and evaluation, the images are downsampled by a factor of 4, resulting in a resolution of 300 × 400 for each image.

\paragraph{Blender.} The Blender \cite{NeRF} dataset consists of 8 synthetic scenes. For fair comparison, we follow previous works \cite{regnerf, freenerf} and use 8 images as input views for each scene, with the following IDs: 26, 86, 2, 55, 75, 93, 16, 73, 8. The 25 test views are sampled evenly from the testing images for evaluation. During training and evaluation, the images are downsampled by a factor of 2, resulting in a resolution of 400 × 400 for each image.

%%%%%%%%%%%%%%%%%%%%%%%%%%%%%%%%%%%%%%%%%%%%%%%%%%%%%%%%%%%%

\end{document}